\documentclass[runningheads]{llncs}
\usepackage{hyperref}
\usepackage{todonotes}
\usepackage[english]{babel}
\usepackage{tikz-uml}
\usepackage{listings}
\usepackage{xcolor}
\usepackage{float}
\usepackage{comment}
\colorlet{punct}{red!60!black}
\definecolor{background}{HTML}{EEEEEE}
\definecolor{delim}{RGB}{20,105,176}
\colorlet{numb}{magenta!60!black}
\newcommand{\casper}[1]{\todo[inline, color=red!30]{Casper: #1}}

\usepackage{subfig}
\usepackage{multirow}
\usepackage{booktabs}
\usepackage{pgfplots}
\usepackage{pgfplotstable}
\usepgfplotslibrary{external}
\makeatletter
\renewcommand{\todo}[2][]{\tikzexternaldisable\@todo[#1]{#2}\tikzexternalenable}
\makeatother
\usepackage[font=small,labelfont=bf,tableposition=top]{caption}
\usepgfplotslibrary{fillbetween}
\usetikzlibrary{calc}
\usepgfplotslibrary{groupplots}
\pgfplotsset{compat=newest}
\lstdefinelanguage{json}{
    basicstyle=\normalfont\ttfamily,
    numbers=left,
    numberstyle=\scriptsize,
    stepnumber=1,
    numbersep=8pt,
    showstringspaces=false,
    breaklines=true,
    frame=lines,
    backgroundcolor=\color{background},
    literate=
     *{0}{{{\color{numb}0}}}{1}
      {1}{{{\color{numb}1}}}{1}
      {2}{{{\color{numb}2}}}{1}
      {3}{{{\color{numb}3}}}{1}
      {4}{{{\color{numb}4}}}{1}
      {5}{{{\color{numb}5}}}{1}
      {6}{{{\color{numb}6}}}{1}t
      {7}{{{\color{numb}7}}}{1}
      {8}{{{\color{numb}8}}}{1}
      {9}{{{\color{numb}9}}}{1}
      {:}{{{\color{punct}{:}}}}{1}
      {,}{{{\color{punct}{,}}}}{1}
      {\{}{{{\color{delim}{\{}}}}{1}
      {\}}{{{\color{delim}{\}}}}}{1}
      {[}{{{\color{delim}{[}}}}{1}
      {]}{{{\color{delim}{]}}}}{1},
}

\newcommand*{\TECHREP}{}%
\ifdefined\PAPER
\title{RMQFMU: Bridging the Real World with Co-simulation\\
  \large For Practitioners\thanks{We are grateful to the Poul Due Jensen Foundation, which has supported the establishment of a new Centre
for Digital Twin Technology at Aarhus University, which will take forward the principles, tools and
applications of the engineering of digital twins.}}
\fi

\ifdefined\TECHREP
\title{RMQFMU: Bridging the Real World with Co-simulation\\
  \large Technical Report\thanks{We are grateful to the Poul Due Jensen Foundation, which has supported the establishment of a new Centre
for Digital Twin Technology at Aarhus University, which will take forward the principles, tools and
applications of the engineering of digital twins.}}
\fi
\author{Mirgita, Henrik, Casper, Lukas}
\titlerunning{RMQFMU: For Practitioners}

\author{Mirgita Frasheri \inst{1}\orcidID{0000-0001-7852-4582} \and
Henrik Ejersbo\inst{1}\orcidID{0000-0003-4753-3800} \and
Casper Thule\inst{1}\orcidID{0000-0001-6606-9236}\and Lukas Esterle\inst{1}\orcidID{0000-0002-0248-1552}}
\authorrunning{M. Frasheri et al.}
\institute{Centre for Digital Twins, DIGIT, Aarhus University \\ 
Finlandsgade 22, 
8200 
Aarhus N, 
Denmark \\
\email{\{mirgita.frasheri,hejersbo,casper.thule,lukas.esterle\}@ece.au.dk}\\
\url{https://digit.au.dk/centre-for-digital-twins/} }
\begin{document}
%\ifdefined\PAPER
%    DEBUG was on
%\fi
%\ifdefined\TECHREP
%TECHREP
%\fi
\maketitle
\begin{abstract}
%\gita{Working on Paper}
    In this paper we present an experience report for the RMQ\-FMU, a plug and play tool, that enables feeding data to/from an FMI2-based co-simulation environment based on the AMQP protocol. 
    Bridging the co-simulation to an external environment allows on one side to feed historical data to the co-simulation, serving different purposes, such as visualisation and/or data analysis. 
    On the other side, such a tool facilitates the realisation of the digital twin concept by coupling co-simulation and hardware/robots close to real-time.
    In the paper we present limitations of the initial version of the RMQFMU with respect to the capability of bridging co-simulation with the real world. To provide the desired functionality of the tool, we present in a step-by-step fashion how these limitations, and subsequent limitations, are alleviated. We perform various experiments in order to give reason to the modifications carried out.
    Finally, we report on two case-studies where we have adopted the RMQFMU, and provide guidelines meant to aid practitioners in its use.

\end{abstract}

%\textbf{Potential Venue:} acsos (early May), \textbf{ieee mrs https://mrs2021.org/ (may)}, ICRA (october)

\ifdefined\PAPER
\section{Introduction}\label{sec:intropap}

Cyber-physical systems (CPSs) refer to systems that combine computational and physical processes, and play an important role in the development of intelligent systems. 
Using real world operation data could possibly facilitate smart decision-making~\cite{banerjee2012ensuring}.
Harnessing this potential is by no means trivial, with challenges including safety~\cite{baheti2011cyber}, reliability, and security~\cite{ecsel2016multi}, among others, that need to be tackled.
Additionally, CPS development is a multi-disciplinary process, implying that different components will be modelled and validated by different tools used in the different involved disciplines. 
The evaluation of such complex systems can be performed through co-simulation~\cite{Gomes&18}. 
%The individual models are integrated into a multi-model, and the joint behaviour of the different components can thereafter be simulated in tandem. 
This is made possible by developing the components according to some standard, e.g. the Functional Mock-up Interface (FMI) adopted in our work, with components referred to as functional mock-up units (FMUs).
However, this is not sufficient, as CPSs together with their environment(s) are subject to continuous changes, and evolve through time, possibly diverging significantly from the initial co-simulated results~\cite{fitzgerald2019multi}.
Digital twins (DTs), defined as digital replicas of the physical components, also known as physical twins (PTs), can be used to follow the behaviour of the PTs and CPSs during operation, potentially through learning~\cite{Fitzgerald&14f}, by adapting the models, as well as performing different tasks such as monitoring or predictions.
%Monitoring deployed cyber-physical systems (CPSs) during operation can assist in detecting undesirable behaviour that can manifest in certain circumstances, as a result of internal failures, or environmental changes. 
The DT and the aforementioned (or more) operations can be developed via the co-simulation of a modelled system that corresponds to the deployed CPS. 
In order to connect the deployed CPS to the DT, the implementation of data brokering between them becomes a necessity.
Such data brokering can be useful in different scenarios; in this paper we identify four such:
\begin{enumerate}
    \item For data analysis and visualisation of system behaviour, where the user is interested in feeding back to the co-simulation environment data recorded from the operation of the hardware, i.e. historical data, in order to perform some analysis on the data, potentially supporting the visualisation of said data.

    \item For co-simulation, where the user is interested in coupling systems that are simulated in different environments.%, and described using different formalisms. \casper{The described in different formalisms seems unnecessary}

    \item For the realisation of the digital shadow (DS), where the digital models of the physical twin are contained within the co-simulation environment.
The user is interested in getting the live data from the PT into the co-simulation environment in order to estimate the difference between actual and simulated data along with visualising the operation of the physical twin.
This is similar to the first usage, the main difference being that here the digital shadow is operating with live data.

    \item For the realisation of the digital twin, where the user is interested in enabling the communication link from the DS to the PT, thus fully realising the implementation of the DT concept.
This means that there are components within the co-simulation environment that, based on the live data, are able to send live feedback to the PT.
This can be useful when we are monitoring the behaviour of the PT, or during the development of CPSs, to ease and lower the cost of testing with Hardware in the Loop.%\casper{Seems like a robot snuck in here}

\end{enumerate}

In this publication we extend an existing Data-Broker called RabbitMQ FMU (RMQFMU)~\cite{thule2020formally}, initially suitable for the first scenario, to enable data brokering both to and from an FMI2-enabled co-simulation, applicable for all the aforementioned scenarios. 
We present an experience report on applying the RMQFMU to two actual cases undergoing development to become digital twin systems: a tabletop robot arm and an autonomous agriculture robot. 
As part of this experience report we present guidelines on how to configure the RMQFMU parameters by presenting various experiments that show their effects and relation to one another.
Through the application of RMQFMU to the aforementioned cases and experiments several needs were discovered such as:
\begin{enumerate}
    \item Data Platform - Get all available data instead of minimally-needed 
    data to enable decision making, possibly allowing to jump ahead to 
    future data. %i.e. the possibility of skipping data to enable larger time 
    %steps.
    \item Performance - RMQFMU shall be as fast as possible.
    \item Data Delay - RMQFMU shall output the newest data if available.
\end{enumerate}

To address need (1) and (2), the existing RMQFMU, henceforth referred to as 
RMQFMU\textsubscript{0}, was realised with multi-threading instead of 
single-threading. This new multi-threaded version is referred to as 
RMQFMU\textsubscript{1}. Experimentation related to detailing the 
configuration of RMQFMU\textsubscript{1} and its related effects led to the 
discovery of need (3).
In order to mitigate (3), yet another version of RMQFMU was realised 
referred to as RMQFMU\textsubscript{2}, which represents the latest version 
of the RMQFMU. %More information is presented in section XX to XY.

The rest of this paper is organised as follows.
The next section provides an overview of the FMI standard, the first version of the RMQFMU (RMQFMU\textsubscript{0}), as well as some of the technologies employed in the realisation of the case-studies.
Afterwards, we informally derive the requirements for RMQFMU, based on the usage scenarios of interest.
Section~\ref{sec:rabbitmq_fmu} presents RMQFMU\textsubscript{1,2}, how it mitigates some inadequacies of RMQFMU\textsubscript{0}, and how it compares to RMQFMU\textsubscript{0}. 
Thereafter, in Section~\ref{sec:experiments} the individual case-studies are described, and the results gained with RMQFMU\textsubscript{1,2} are presented.
Moreover, a detailed comparison between RMQFMU\textsubscript{1} and RMQFMU\textsubscript{2} is made. 
Section~\ref{sec:guidelines} provides a set of guidelines meant to aid the use of RMQFMU\textsubscript{2}.
Concluding remarks and potential avenus for future work are presented in Section~\ref{sec:conclusion}.

\begin{comment}
\vspace{5em}
OLD COMMENTS (Before March 8, 2021)
\begin{itemize}
\item talk about the digital twin (DT) concept.

\item give a motivation for the rabbitMQ fmu, in terms of its role in connecting the co-simulation and the real world, connecting to the previous paragraphs on the DT.
\item contributions:
\begin{itemize}
\item talk about the rabbitMQ fmu - high level description, and main requirement
\item performance evaluation of alternative designs.
\end{itemize}
\end{itemize}
\end{comment}

\fi

\ifdefined\TECHREP
\section{Introduction}\label{sec:introtechrep}

Cyber-physical systems (CPSs) refer to systems that combine computational and physical processes, and play an important role in the development of intelligent systems. 
Using real world operation data could possibly facilitate smart decision-making~\cite{banerjee2012ensuring}.
Harnessing this potential is by no means trivial, with challenges including safety~\cite{baheti2011cyber}, reliability, and security~\cite{ecsel2016multi}, among others, that need to be tackled.
Additionally, CPS development is a multi-disciplinary process, implying that different components will be modelled and validated by different tools used in the different involved disciplines. 
The evaluation of such complex systems can be performed through co-simulation~\cite{Gomes&18}. The individual models are integrated into a multi-model, and the joint behaviour of the different components can thereafter be simulated in tandem. 
However, this is not sufficient, as CPSs together with their environment(s) are subject to continuous changes, and evolve through time, possibly diverging significantly from the initial co-simulated results~\cite{fitzgerald2019multi}.
Digital twins (DTs), defined as digital replicas of the physical components, also known as physical twins (PTs), can be used to follow the behaviour of the PTs and CPSs during operation, potentially through learning~\cite{Fitzgerald&14f}, by adapting the models, as well as performing different tasks such as monitoring or predictions.
%Monitoring deployed cyber-physical systems (CPSs) during operation can assist in detecting undesirable behaviour that can manifest in certain circumstances, as a result of internal failures, or environmental changes. 
The DT and the aforementioned (or more) operations can be developed via the co-simulation of a modelled system that corresponds to the deployed CPS. 
In order to connect the deployed CPS to the DT, implementing data brokering between them becomes a necessity.

The work presented in this publication extends an existing Data-Broker called RabbitMQ FMU (RMQFMU)~\cite{thule2020formally}, to enable data brokering to and from an FMI2-enabled co-simulation. 
We present an experience report on applying RMQFMU to two actual cases undergoing development to become digital twin systems: an autonomous agriculture robot and a tabletop robot arm. As part of this experience report we present guidelines on how to configure the RMQFMU parameters by presenting various experiments that show their effects and relation to one another.
Through the application of RMQFMU to the aforementioned cases and experiments several needs were discovered such as:
\begin{enumerate}
    \item Data Platform - Get all available data instead of minimally-needed data to enable decision making - i.e. the possibility of skipping data to enable larger time steps.
    \item Performance - RMQFMU shall be as fast as possible.
    \item Data Delay - RMQFMU shall output the newest data if available.
\end{enumerate}

To address need 1. and 2., the existing RMQFMU, henceforth referred to as RMQFMU\_0, was realised with multi-threading instead of single-threading. This new multi-threaded version is referred to as RMQFMU\_1. Experimentation related to detailing the configuration of RMQFMU\_1 and its related effects led to the discovery of need 3.
In order to mitigate 3. yet another version of RMQFMU was realised referred to as RMQFMU\_2. More information is presented in section XX to XY.

The rest of this paper is organised as follows.
The next section provides an overview of the FMI standard, the first version of the RMQFMU (RMQFMU\textsubscript{0}), as well as some of the technologies employed in the realisation of the case-studies.
In Section~\ref{sec:motivation}, we describe four scenarios with which we motivate the need for the RMQFMU.
Afterwards, we informally derive the requirements for RMQFMU, based on the usage scenarios of interest.
Section~\ref{sec:rabbitmq_fmu} presents RMQFMU\textsubscript{1,2}, how it mitigates some inadequacies of RMQFMU\textsubscript{0}, and how it compares to RMQFMU\textsubscript{0}. 
Thereafter, in Section~\ref{sec:experiments} the individual case-studies are described, and the results gained with RMQFMU\textsubscript{1,2} are presented.
Moreover, a detailed comparison between RMQFMU\textsubscript{1} and RMQFMU\textsubscript{2} is made. 
Section~\ref{sec:guidelines} provides a set of guidelines meant to aid the use of RMQFMU\textsubscript{2}.
Concluding remarks and potential avenus for future work are presented in Section~\ref{sec:conclusion}.

%\lukas{introduction digital/physical twin}

\begin{comment}
\vspace{5em}
OLD COMMENTS (Before March 8, 2021)
\begin{itemize}
\item talk about the digital twin (DT) concept.

\item give a motivation for the rabbitMQ fmu, in terms of its role in connecting the co-simulation and the real world, connecting to the previous paragraphs on the DT.
\item contributions:
\begin{itemize}
\item talk about the rabbitMQ fmu - high level description, and main requirement
\item performance evaluation of alternative designs.
\end{itemize}
\end{itemize}
\end{comment}

\fi

\section{Background}\label{sec:background}

In this Section we provide a brief summary on the relevant concepts for this work, such as co-simulation, FMI standard, master algorithms, and the tools we use for their realisation. 
Thereafter, we present the first released version of the RMQFMU (RMQFMU\textsubscript{0}), which we have extended as presented in this paper. 
\subsection{Concepts and Tools}
The realisation of CPSs and constituent systems is a cross-disciplinary process, where the different components are modelled using different formalisms and modelling tools~\cite{gomes2018co}. 
In order to evaluate the behaviour of such systems as a whole, co-simulation techniques are used, which require the integration of the separate models into what are called multi-models~\cite{7496424}.
For the latter to be possible, the tools used to produce the individual models need to adhere to some standard. 
One such standard is the Functional Mock-up Interface 2.0 for Co-simulation (FMI)~\cite{FMIStandard2.0.1} that defines the C-interfaces to be exposed by each model, as well as interaction constraints, packaging, and a static description format. 
An individual component that implements the FMI standard is called Functional Mock-up Unit (FMU).
A co-simulation is executed by an orchestration engine that employs a given master algorithm. The master algorithm defines the progression of a co-simulation in terms of getting outputs via \texttt{getXXX} function calls, setting inputs via \texttt{setXXX} function calls, and stepping the individual FMUs in time via the \texttt{doStep} function calls.
Common master algorithms are the Gauss-Seidel and Jacobi (the curious reader is referred to~\cite{gomes2018co} for an overview of these algorithms and a survey on co-simulation).
In this work, we use the open-source INTO-CPS tool-chain~\cite{Fitzgerald&15}, for the design and execution of co-simulation multi-models, with Maestro~\cite{Thule&19} as orchestration engine employing an FMI-based Jacobi master algorithm. %\casper{Do we even usen Maestro?}
%\gita{In the first case-study, I run the experiments with the intocps and maestro, so yes.}
%\subsection{Case-Studies} \gita{Should this be here at all? Case-studies are described later on}
%\gita{not here, we will mention in the intro that we have, then there is the dedicated section for them.}
\subsection{Overview of the RMQFMU\textsubscript{0}}
%\gita{We can add the diagram flows here maybe? Or would it take too much space?}
%\gita{we need to add code url as footnote. Casper: Done}
%\gita{Should we describe more the water-tank case-study. Casper: No}
RMQFMU\textsubscript{0} was implemented to enable getting external/historical data into the co-simulation environment, for either replaying such data in simulation, or for performing different kinds of analysis, e.g. checking its difference from expected data (the reader is referred to the Water-Tank Case-Study for an example\footnote{Available at \url{https://github.com/INTO-CPS-Association/example-single_watertank_rabbitmq}, visited May 6, 2021}). 
Essentially, RMQFMU\textsubscript{0} subscribes to messages with a specified routing key and outputs messages as regular FMU outputs at specific points in time based on the timestamp of the messages.
At every call of the \texttt{doStep} function, the RMQFMU\textsubscript{0} attempts to consume a message from the server.
A retrieved message is placed in an incoming queue, from which it is thereafter processed, according to the quality constraints (\textit{maxage} and \textit{lookahead}), described shortly.
%\gita{we're in a tough spot here, given that there will not be more than one message at a time in this queue for this version... how to put it in a nice way?}
In case there is no data available, the RMQFMU\textsubscript{0} will wait for a configurable timeout before exiting.
Note that, the \texttt{doStep} is tightly bound with the consuming operation, i.e. they are both contained within the main thread of execution. 
If on the other hand there is data, its validity with respect to time is checked.
The \textit{maxage}, main parameter of the RMQFMU\textsubscript{0}, allows to configure the age of data within which the data is considered valid at a given time-step.
Additionally, the \textit{lookahead} parameter specifies how many messages will be processed at a step of RMQFMU and thereby at the given point in time\footnote{In this version of the RMQFMU the functionality based on the \textit{lookahead} is not implemented fully.}. 
%The main parameter of the RBMQFMU\_0 is the \textit{maxage}, which constrains the validity of a received message based on its timestamp.
%The RBMQFMU\_0's main parameters are the \textit{maxage} and \textit{lookahead}, where the \textit{maxage} constrains the validity of a received message based on its timestamp, and the \textit{lookahead} specifies how many messages will be processed at a step of RMQFMU and thereby at the given point in time. 
The behaviour of the RMQFMU\textsubscript{0} has been formally verified in previous work~\cite{thule2020formally}.

\ifdefined\TECHREP
\section{Motivation for the RMQFMU}\label{sec:motivation}
%\casper{Shorten the entire motivation and move to introduction?}
%\casper{Skip the INTO-CPS stuff here. The RMQFMU is generic in nature }
The purpose of the RMQFMU is to serve as a bridge between an FMI-based co-simulation environment, and an external component, e.g. a simulator or hardware, that needs to be coupled with the co-simulation environment in order to achieve a predefined goal.
We consider four scenarios in which such FMU is useful.

\paragraph{Data analysis and visualisation of system behaviour}
In this case, the user of the RMQFMU is interested in feeding back to the co-simulation environment data recorded from the operation of the hardware, i.e. historical data, in order to perform some analysis on the data, potentially supporting the visualisation of said data, e.g. Water-Tank Case-study referred to in Section~\ref{sec:background}.
%\casper{We need to introduce the water-tank case study if we plan to use it like this}

\paragraph{Co-simulation}
In this scenario the user of the RMQFMU is interested in running a co-simulation of a given system.
Some of the constituents of the system can behave as FMUs, while others cannot, e.g. a robot simulation in Gazebo, and therefore has to be connected to the co-simulation by other means.

\paragraph{Realisation of the Digital Shadow}
In this scenario, digital models of the physical twin are contained within the co-simulation environment.
The user is interested in getting the live data from the PT into the co-simulation environment in order to estimate the difference between actual and simulated data along with visualising the operation of the physical twin.
This is similar to the first usage, the main difference being that here the digital shadow (DS) is operating with live data.

\paragraph{Realisation of the Digital Twin}
In this scenario, the user is interested in enabling the communication link from the DS to the PT, thus fully realising the implementation of the DT concept.
This means that there are components within the co-simulation environment that, based on the live data, are able to send live feedback to the PT.

This can be useful when we are monitoring the behaviour of the PT, e.g. consider a robot moving from point A to B.
Assume an FMU that models the motion of the robot, and can be used to predict where the robot should be after some time given the starting point and velocity.
As the live data of the robot's position comes in, the divergence between the predicted and actual paths can be used to trigger an emergency stop to the robot~\footnote{This is considered a supervisory control and not the main safety mechanism.}.
This setup can also be useful during the development of the robot, to ease and lower the cost of testing with Hardware in the Loop.
%Assume the user wants to inject faults in the system and evaluate the fault tolerance mechanisms in place. 
%As an example, the user might want to tamper with the controller of the robot. 
%To explore this scenario, the user can place a controller within the Digital Twin, to which faulty values can be injected, that lead to undesired behaviour of the robot, e.g. going too close to the obstacle. 
%The advantage here is that the user would not need to touch the native controller.\gita{Is the aforementioned a reasonable argument?}
%The safety monitor mechanism presented in the previous example can be tested in the present scenario.

\fi

\section{The RMQFMU}\label{sec:rabbitmq_fmu}
%This section provides an overview of the requirements that underlie the development of the RMQFMU, and description of the system design that has been implemented.
The design of the RMQFMU is based on a set of informally derived requirements, covering those functionalities needed for the adoption of the RMQFMU in the scenarios described in
\ifdefined\PAPER
Section~\ref{sec:intropap}.
\fi
\ifdefined\TECHREP
Section~\ref{sec:motivation}.
\fi
These requirements are given as follows:
%\casper{Mention that this is in no way a full requirement analysis or anything like it}

\begin{enumerate}
    \item The RMQFMU is able to get the data published by an external system to the RabbitMQ server. 
    This is relevant for all scenarios.
    \item The RMQFMU is able to publish data to the RabbitMQ server, thus closing the communication link. 
    This is relevant for the second and fourth scenario, i.e. for co-simulation and DT realisation.%, in which it is needed to send data to the external system.
    
    %%FUTURE FEATURE
    %\item The FMU is able to output information on the state of its buffer, e.g. whether it contains future data.
    %on the time difference between its current step, and the time of the external component.
    %This is relevant for the second and fourth scenario, in which the synchronisation between systems becomes important.
    %%FUTURE FEATURE
    %\item The FMU supports configuration with respect to how data is handled, e.g.:
    %\begin{itemize}
    %    \item do not skip over data, 
    %    \item jump to latest data,
    %    \item do not skip over tagged data.
    %\end{itemize}
    %This is relevant for the second and fourth scenario. Consider a case where the messages arrive in bursts in the DT. As the DT is processing older messages, it is behind in time with respect to its physical twin. In order to get back in sync, it might be necessary to jump directly to the latest messages in the queue. 
    %Note that, the strategy to be adopted here depends on how data is processed by the other entities in the DT, e.g. monitors. Indeed for some monitors skipping to the latest data is not a viable strategy, as all the historical data might be needed to ensure correct behaviour. 

    \item The FMU steps as fast as possible. Indeed the role of the RMQFMU is that of a data broker, as such it is not part of the system being simulated, rather a facilitating entity. 
    Therefore, the delays it causes should be minimal and impact the overall co-simulation environment as little as possible.
    \item Provide quality constraints, e.g. with respect to age of data (\textit{maxage}).
    \item Provide performance constraints, e.g. number of messages processed per step (\textit{lookahead}).
    
\end{enumerate}
%\gita{we need a transition sentence here to connect to what comes next, and say that some of these we have tackled, like maxage and lookahead, but we need to do the rest.}
% \gita{Fixed, please check.}
%  Requirements 1 and 4 are already fulfilled with RMQFMU\textsubscript{0}.
 RMQFMU\textsubscript{0} already fulfils requirements 1 and 4.
 Requirement 5 is partially implemented in this version as well, in that while the \textit{lookahead} determines how many messages to retrieve from the incoming queue at every time-step, there will not be more than one message at a time in this queue, given that a consume call is performed at the \texttt{doStep} call until a valid message is retrieved. 
In order to tackle requirement 2, we enable the configuration of the inputs of the RMQFMU\textsubscript{1,2} as needed, i.e., a user can define as many inputs as desired, of one the following types: integer, double, boolean, and string. 
These inputs will be sent to the RabbitMQ server, on change; in other words, the values of the inputs will be forwarded if they have changed as compared to the previous step taken by the RMQFMU\textsubscript{1,2}. 
This check is performed within the \texttt{doStep}, at every time-step. 

In order to tackle requirement 3, we investigate the potential benefit of a threaded configuration of the RMQFMU\textsubscript{1,2}. 
Indeed, the FMU allows for a build-time option to enable a multi-threaded implementation, with a separate thread to interface and consume data from the RabbitMQ server, parsing incoming data and placing it in the FMU incoming queue. 
The main \texttt{doStep} function of the FMU still executes in the main thread context of the calling simulation orchestration application, and it reads and processes data from the FMU incoming queue and produces outputs. If the multi-threaded option is disabled (default), the \texttt{doStep} function will consume data directly from the RabbitMQ server in the context of the calling main thread. %\casper{Maybe the incoming queue stuff should go to introduction of initial version? }\gita{mentioned it there, but how to put it in a nice way the functionality is not full, wrt lookahead, and the fact that we have one message in the queue at a time no matter what.}

In general the potential performance benefits of the threaded implementation will depend on the amount of data being consumed, parsed, and provided in the internal incoming FMU queue when the \texttt{doStep} function is called. 
Hence, the benefits will depend largely on the co-simulation environment. 
If the co-simulation is fast, in vague terms if the delay between each call of the FMU \texttt{doStep} function is short compared to the rate of the incoming data, then the FMU will be mostly blocking for I/O. 
In this situation, the separate consumer thread may not provide much benefit as the \texttt{doStep} function will still need to wait/block for incoming data.
On the other hand, if the co-simulation is slower compared to the rate of data, then the separate thread may be able to consume, parse, and provide data to the internal incoming FMU queue, in parallel to the main co-simulation thread running its orchestration engine and executing other potential FMUs or monitors. 
In addition, any performance benefits of a threaded implementation also depends largely on the execution platform having a multi-core processor and the OS/thread environment to be able to take advantage.

The threaded implementation allows for the full realisation of the \textit{lookahead} functionality, thus tackling requirement 5. 
The consumer thread is continuously retrieving data -- when available -- and placing these messages in the incoming queue, from where said data is processed in chunks of \textit{lookahead} size. 
%To enable testing with a varying delay impact of the co-simulation environment, the testing of this case-study is performed in a setting with a mock-up test environment. This test environment allows configuration of the delay imposed by the co-simulation, basically by allowing the delay between calls to the RabbitMQ \texttt{doStep} function to be controlled by the test.

\section{Experiments}~\label{sec:experiments}
The RMQFMU\textsubscript{1,2} has been evaluated with a series of experiments, across different combinations of parameters such as the maxage and lookahead, and in two different case-studies. %\casper{lookahead not introduced}
The purpose of these setups is to provide an understanding on the performance of the RMQFMU\textsubscript{1,2}, parameter tuning, as well as provide some insight into the effect of external factors.
The first case-study presents a scenario based on data from a deployed industrial robot, thus adequately representing an industrial case. 
As a result, we perform the performance evaluation in this case-study. 
Additionally, we add mockup components to the co-simulation structure, to mimic a more realistic co-simulation with different components for different purposes, such as data analysis, prediction among others. 
%The mockup setup allows us to investigate in-depth the benefits of the threaded option for the RMQFMU.
Additionally, we consider the impact of external factors, such as the frequency of sending data to the RMQFMU\textsubscript{1,2}. %, and the possibility of additional RMQFMUs coupled in the same co-simulation.
Whereas, the second case-study presents a setup with the Gazebo simulation of the Robotti agricultural robot~\cite{foldager2018design}, rather simple both in terms of the size of the messages being sent and the internal structure of the co-simulation.
The parameters of interest are the \textit{lookahead}, \textit{maxage}, and the frequency with which we send data into the RMQFMU\textsubscript{1,2}.

The experimentation is divided in two phases. 
In the first, we evaluate the impact of the multi-threaded implementation on the performance of the RMQFMU.
Specifically, we compare RMQFMU\textsubscript{1} to RMQFMU\textsubscript{0}, and argue for the suitability of adopting the multi-threaded configuration as the primary implementation for the RMQFMU.
Additionally we consider the effects of the maxage and lookahead for RMQFMU\textsubscript{1}.
In the second phase, we evaluate the alterations to RMQFMU\textsubscript{1}, i.e. RMQFMU\textsubscript{2}, in order to tackle the side effects observed in phase 1, providing both performance and behaviour results.

\subsection{Case-study 1: Universal robot 5e (UR)}
This case study concerns an industrial robot called UR5e\footnote{Available at \url{https://www.universal-robots.com/products/ur5-robot/}, visited April 7, 2021} (Figure~\ref{fig:ur5e}). 
The robot has a reach of 85 mm, payload of 5kg and 6 rotating joints. 
It is chosen because the UR5e is in production, thus represents a realistic component in a digital twin setting, and thereby a realistic amount of data. 
The data set is generated from a series of movements used for calibrating the digital twin to a related physical twin. 
Therefore, all joints are exercised. 
%The data set is in csv format and consists of 35123 lines (including header), each (except header) with 107 float values (one being time) and 10 integer values.
The test data from the UR robot used is present in csv format (35123 lines including the header) and consists of messages (excluding the header) with 107 float values (one being time) and 10 integer values. 
The messages have a sample rate (frequency $f_{data}$) of $2$ms and will be replayed/produced into the RabbitMQ broker with this rate. 
The test data contains a substantial number of time gaps, i.e.\ places where two consecutive messages in the input data are spaced with more than $2$ms. 
Data is replayed at a constant frequency of $2$ms and as such some messages are produced faster than their timestamps indicate. The effects of this will be discussed in the test results. 
Testsare also performed on a \textit{cleaned} test data with a constant message spacing of $2$ms between all pairs of messages - i.e.\ no additional time gaps.

To investigate the behaviour of the FMU, we perform tests with different variations and relations between the simulation step size and the delay imposed by the simulation environment. 
We perform the tests with and without the threading option to investigate its effect, $t_{\mathit{on}}$ and $t_{\mathit{off}}$ respectively. 
We first consider simulation step duration and environment delays that simulates a setup where data is expected to be consumed as fast as its produced - i.e.\ every $2$ms and the overhead of the simulation delay is minimal. 
In this case we consider both the simulation step and the simulation delay to match the input data rate of $2$ms. 
Then we consider situations with larger simulation step duration and larger simulation delays - to mimic co-simulation environments with additional FMUs or monitors. 
The full list of test data used is provided in Table~\ref{tab:exp1}.
In short, the results will provide an insight into how different parameters like frequency of input data, simulation step size, and simulation delay, and the relation of these parameters, influence the performance of the RMQFMU.
%\casper{This has a wrong reference for some reason} 
%\gita{label has to be put directly after \\caption, then it's fine}
Note, that for these tests we use a fixed \textit{lookahead} of $1$ and a fixed \textit{maxage} of $300$ms. The max age is chosen as such to cover for time gaps in the input data up to this range.

\begin{minipage}{\textwidth}
  \begin{minipage}[t]{0.4\textwidth}
    \centering
    \raisebox{-15ex-\height}{\includegraphics[scale=0.15]{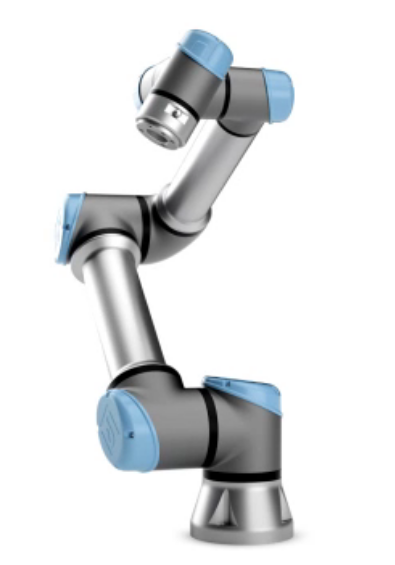}}
    \captionof{figure}{UR5e robot%~\protect\footnotemark
    }
    \label{fig:ur5e}
  \end{minipage}
  \hspace{0.5cm}
  \begin{minipage}[t]{0.4\textwidth}
    \centering
     \begin{table}[H]
\centering  
\caption{Tests for Case-study $1$~\protect\footnotemark}\label{tab:exp1} 
\begin{tabular}{cccccccc}

\multirow{2}{*}{} & \multicolumn{3}{c}{}                                                                                                                    \\
\toprule                                                                        
            & {Sim step} & {Sim delay} & {Thread} \\
\midrule
Case 1      & 2ms  & 2ms & $t_{\mathit{off}}$ \\
\midrule
Case 2      & 2ms  & 2ms & $t_{\mathit{on}}$  \\
\midrule
Case 3$^*$  & 100ms & 100ms & $t_{\mathit{off}}$ \\
\midrule                                                                                                 
Case 4$^*$  & 100ms & 100ms & $t_{\mathit{on}}$  \\
\midrule
Case 5      & 100ms & 113ms & $t_{\mathit{on}}$  \\
\midrule                                                                                                 
Case 6      & 100ms & 120ms & $t_{\mathit{on}}$  \\
\bottomrule
\end{tabular}
%ç
\end{table}

    \end{minipage}
  \end{minipage}
  \ifdefined\PAPER
 \footnotetext{Due to space constraints, we present in the paper results for some selected cases marked with $*$. The results as a whole can be found in the technical report\cite{frasheri2021rmqfmu}.}
 \fi
%\footnotetext{(\url{https://www.universal-robots.com/products/ur5-robot/}, visited April 7, 2021)} 

%The purpose of this case-study is to provide an insight into the performance of the RMQFMU -- specifically the benefit of the multi-threaded approach -- when used to handle real-world input data from an industrial robot. 
%The sample robot data has a large number of values per message (117) and messages represent data being sampled with an interval of $2$ms. 
%We will in particular be considering the potential effects of using a multi-threaded configuration of the RMQFMU to improve performance. 
%The test results will provide an insight into how different parameters like frequency of input data, simulation step size, and simulation delay, and the relation of these parameters, influence the performance of the RMQFMU. %And as such are important to consider when configuring a co-simulation setup using the RMQFMU.

%In Section~\ref{sec:rbmq_fmu_multi_thread_setup} we describe in more detail, the setup of using the RMQFMU in a multi-threaded configuration. In Section~\ref{sec:rbmq_fmu_test_data} we motivate and present the test cases and test data used for this case study. And in Section~\ref{sec:rbmq_fmu_test_results} we present and discuss the test results for the different tests.

\subsection{Case-study 2: Gazebo Simulation of Robotti}
The purpose of this case-study is to provide a basic example of a co-simulation coupled with the Gazebo simulation of the Robotti agricultural robot~\cite{foldager2018design}, through the RMQFMU, in order to gain insight into how the configurable parameters of the RMQFMU affect its behaviour.
In this case the co-simulation environment consists of the RMQFMU and a monitor FMU. 
%In this case the Gazebo simulation of the robot is coupled to the co-simulation environment consisting of the safety monitor and RMQFMU, through the latter.  
The data of interest in this scenario, i.e. the data sent through the RMQFMU consists of: the $x$ and $y$ positions of the robot, and the $x$ and $y$ position of the nearest obstacle. 
Additionally, a sequence number is attached to each message to keep track of the outputted messages during the processing of the results.
The monitor takes as input such data, and computes the distance between the robot and obstacle for every time-step.
In case such distance is below a predefined safety threshold, the monitor will issue an emergency stop to the Gazebo simulation of the robot.
%On the other side, the robot is sending continuously data regarding its position and that of the nearest obstacle. 

\begin{table}[htb]
\centering
\caption{Overview of experiments for Case-study (CS) 2}\label{tab:exp2}
\begin{tabular}{c | c c c c || c | c c c c || c | c c c c}
\toprule
CS 2 & \multicolumn{4}{c ||}{Parameters}                    & CS 2 & \multicolumn{4}{c ||}{Parameters}   & CS 2 & \multicolumn{4}{c}{Parameters}\\
             & lA  & mA   & $t_s$ & $f_{data}$ &                   & lA   & mA    & $t_s$ & $f_{data}$   &    & lA   & mA    & $t_s$ & $f_{data}$ \\
\midrule
Case 1       & 1   & 0.2s & 0.1   & 10Hz &  Case 5                 & 50  & 0.2s & 0.1   & 500Hz       &  Case 9$^*$             & 2   & 0.4s & 0.1   & 5Hz       \\
\midrule
Case 2       & 50  & 0.2s & 0.1   & 10Hz & Case 6                  & 200 & 2s   & 0.1   & 500Hz     &  Case 10$^*$            & 5   & 0.4s & 0.1   & 5Hz\\
\midrule
Case 3       & 1   & 2s   & 0.1   & 10Hz &  Case 7                 & 50  & 0.2s & 0.1   & 500Hz   &  Case 11$^*$      & 2   & 2s   & 0.1   & 5Hz\\
\midrule
Case 4$^*$   & 50  & 2s   & 0.1   & 10Hz &  Case 8                 & 200 & 2s   & 0.1   & 500Hz   & Case 12$^*$      & 5   & 2s   & 0.1   & 5Hz\\

\bottomrule
\end{tabular}
\end{table}

%\gita{mark the case-studies in the table we provide graphs for. And remove the thread column, just mention in text.}
For this case-study we are interested in the following parameters (see Table~\ref{tab:exp2}), the maxage $mA$ of the messages, and the lookahead $lA$, and how they affect the behaviour of the RMQFMU, i.e. in terms of the sequence of messages outputted at every time-step. 
Note that for all the cases the threaded option is used ($t_{\mathit{on}}$).
Parameters such as the size of the time-step $t_s$, the frequency of sending data $f_{data}$ to the co-simulation are fixed for cases $1-4$, such that they align. 
We argue such decision with the simplicity of the case-study, which could allow a user in reality to align these values. 
The time-step of the simulation is fixed to $0.1$s to provide adequate granularity.
Whereas for cases $5-12$ we consider $f_{data}$ that doesn't match the size of the simulation step, and observe the effects of the maxage and lookahead. 
%Additionally, we configure the RMQFMU with the threaded option enabled as we expect a negligible impact of this parameter in the current setting.
%Nonetheless, the effect of the threaded option is tackled in-depth in the subsequent case-study.

\subsection{Phase 1 Results}
\ifdefined\PAPER
\begin{figure}[t]
  \centerline{{\pgfdeclarelayer{main}
\pgfdeclarelayer{foreground}
\pgfdeclarelayer{background}
\pgfsetlayers{background,main,foreground}
\begin{tikzpicture}
    \pgfplotstableread[col sep=comma]{raw_data/rbmqv1/urdata/ctoff_s100_z100.csv}\datatable
    \pgfplotstableread[col sep=comma]{raw_data/rbmqv1/urdata/cton_s100_z100.csv}\datatabletwo
    
    \pgfplotstablegetrowsof{\datatable}
    \pgfmathsetmacro{\N}{\pgfplotsretval} 

    \begin{groupplot}[group style={
                    group name=myplot,
                    group size= 2 by 1,vertical sep=1.cm},height=3.5cm,width=6cm,
                    legend style={at={(1.1,-.3)},anchor=north,legend columns=-1},
                    xmin=-0.1,
                    xmax=20.1,
                    %minor x tick num = 4,
                    %minor y tick num = 4,
                    %xminorgrids=true,
                    %yminorgrids=true,
                    %xmajorgrids=true,
                    %ymajorgrids=true,
                    scaled y ticks=base 10:-3,
                    ]
        \nextgroupplot[ylabel={Step duration ($\mu s$)},ymax=20000, xlabel={sim-time (s)}]
            \addplot[mark=none, blue] table[x index = {0},y index = {1}] \datatable;
            \addplot[mark=none, dashed, red] table[x index = {0},y index = {1}] \datatabletwo;
            \addlegendentry{{$t_{\mathit{off}}$}}
            \addlegendentry{{$t_{\mathit{on}}$}}
            \addplot [name path=upper,draw=none] table[x index = {0},y expr=\thisrowno{1}+\thisrowno{2}] \datatable;
            \addplot [name path=lower,draw=none] table[x index = {0},y expr=\thisrowno{1}-\thisrowno{2}] \datatable;
            \addplot [fill=blue!20] fill between[of=upper and lower];
            \addplot [name path=upper,draw=none] table[x index = {0},y expr=\thisrowno{1}+\thisrowno{2}] \datatabletwo;
            \addplot [name path=lower,draw=none] table[x index = {0},y expr=\thisrowno{1}-\thisrowno{2}] \datatabletwo;
            \addplot [fill=red!20] fill between[of=upper and lower];

        %\nextgroupplot[ylabel={Sequence no}, xlabel={sim-time}]
        %    \addplot[mark=none, blue] table[x index = {0},y index = {3}] \datatable;
        %    \addplot[mark=none, dashed, red] table[x index = {0},y index = {3}] \datatabletwo;
        %    \addplot [name path=upper,draw=none] table[x index = {0},y expr=\thisrowno{3}+\thisrowno{4}] \datatable;
        %    \addplot [name path=lower,draw=none] table[x index = {0},y expr=\thisrowno{3}-\thisrowno{4}] \datatable;
        %    \addplot [fill=blue!20] fill between[of=upper and lower];
%
        %    \addplot [name path=upper,draw=none] table[x index = {0},y expr=\thisrowno{3}+\thisrowno{4}] \datatabletwo;
        %    \addplot [name path=lower,draw=none] table[x index = {0},y expr=\thisrowno{3}-\thisrowno{4}] \datatabletwo;
        %    \addplot [fill=red!20] fill between[of=upper and lower];
            
        %\coordinate (pt) at (axis cs:21,5600);
        %\draw[->, >=stealth', black!70, dashed] (axis cs:3,50) -- (axis cs:15,2000);

         \nextgroupplot[ylabel={Queue size}, xlabel={sim-time (s)}, ymax=800,scaled y ticks=base 10:-2,]
            \addplot[mark=none, blue] table[x index = {0},y index = {5}] \datatable;
            \addplot[mark=none, dashed, red] table[x index = {0},y index = {5}] \datatabletwo; 
            \addplot [name path=upper,draw=none] table[x index = {0},y expr=\thisrowno{5}+\thisrowno{6}] \datatable;
            \addplot [name path=lower,draw=none] table[x index = {0},y expr=\thisrowno{5}-\thisrowno{6}] \datatable;
            \addplot [fill=blue!20] fill between[of=upper and lower];
            \addplot [name path=upper,draw=none] table[x index = {0},y expr=\thisrowno{5}+\thisrowno{6}] \datatabletwo;
            \addplot [name path=lower,draw=none] table[x index = {0},y expr=\thisrowno{5}-\thisrowno{6}] \datatabletwo;
            \addplot [fill=red!20] fill between[of=upper and lower];

    \end{groupplot}

    % this is the inset plot ...
    %\begin{axis}[
    %	tiny,
        % ... which should be plotted at the stored coordinate ...
    %    at={(pt)},
        % ... with this `anchor'
    %    anchor=north east,
        % use this predefined style (it is predefined by PGFPlots itself)
        %
        % now state the options which should be used for the inset plot
    %    width=2.5cm,
    %    height=2.5cm,
    %    xtick distance=.2,
    %    xmin=0,xmax=.4,
        %xmajorgrids=true,
        %ymajorgrids=true,
        % use this key to fill the background of the axis only
    %    axis background/.style={
    %        fill=white,
    %    },
        % name this axis so it can later be used to fill the "background" of the
        % whole plot including the labels
    %    name=insetAxis,
    %]
%        \addplot [mark=x] table [x=size, y=cluster] {sim-0.3-0.6.csv};
        % again the dummy plot
        
     %   \addplot[mark=none, blue] table[x index = {0},y index = {3},skip coords between index={5}{\N}] \datatable;
     %   \addplot[mark=none, dashed, red] table[x index = {0},y index = {3},skip coords between index={5}{\N}] \datatabletwo;
    %\end{axis}
%
\end{tikzpicture}}}
  \caption{Cases 3 and 4: step=100ms, delay=100ms, 
  thread=$t_{\mathit{off}}$/$t_{\mathit{on}}$ }
  \label{fig:log_f2_s100_z100_orig}
\end{figure}
\fi

\subsubsection{Case-study 1}

\ifdefined\TECHREP
%The results of case 1 (thread off) are shown in Figure~\ref{fig:log_ctoff_f2_s2_z2_orig}.
%\begin{figure}
%  \centerline{\includegraphics[scale=0.60]{Figures/log_ctoff_f2_s2_z2_orig.pdf}}
%  \caption{Case 1: step=2ms, delay=2ms, thread=off}
%  \label{fig:log_ctoff_f2_s2_z2_orig}
%\end{figure}

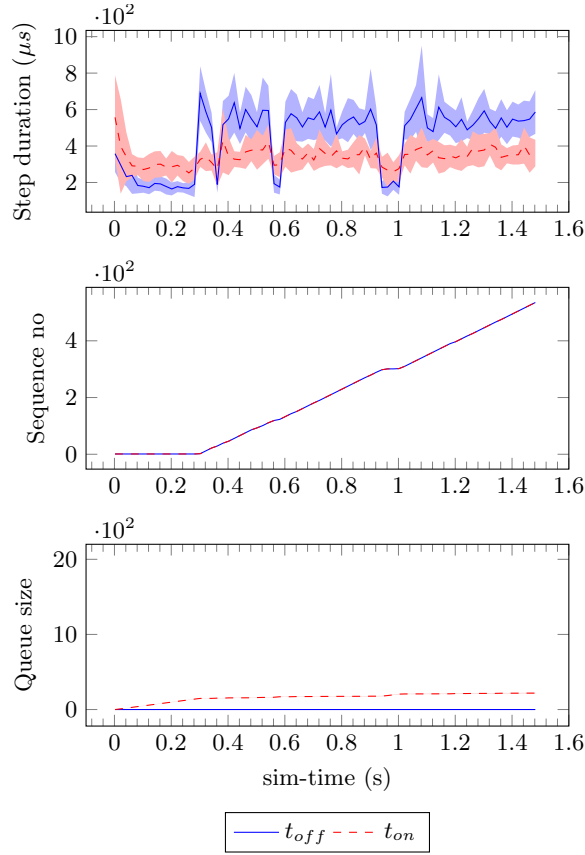
\begin{figure}[tbh!]
  \centerline{{\begin{tikzpicture}
    \pgfplotstableread[col sep=comma]{raw_data/rbmqv1/urdata/ctoff_s2_z2_cut.csv}\datatable
    \pgfplotstableread[col sep=comma]{raw_data/rbmqv1/urdata/cton_s2_z2_cut.csv}\datatabletwo
    \pgfplotstablegetrowsof{\datatable}
    \pgfmathsetmacro{\N}{\pgfplotsretval} 
    \begin{groupplot}[group style={
                    group name=myplot,
                    group size= 1 by 3,vertical sep=1cm},height=4cm,width=8cm,
                    legend style={at={(.5,-3.3)},anchor=north,legend columns=-1},
                    xmin=-0.1,
                    xmax=1.6,
                    minor x tick num = 4,
                    %minor y tick num = 4,
                    %xminorgrids=true,
                    %yminorgrids=true,
                    %xmajorgrids=true,
                    %ymajorgrids=true,
                    scaled y ticks=base 10:-2,
                    ]
        \nextgroupplot[ylabel={Step duration ($\mu s$)}]
            \addplot[mark=none, blue] table[x index = {0},y index = {1},skip coords between index={75}{\N}] \datatable;
            \addplot[mark=none, dashed, red] table[x index = {0},y index = {1},skip coords between index={75}{\N}] \datatabletwo;
            \addlegendentry{{$t_{\mathit{off}}$}}
            \addlegendentry{{$t_{\mathit{on}}$}}
            \addplot [name path=upper,draw=none] table[x index = {0},y expr=\thisrowno{1}+\thisrowno{2},skip coords between index={75}{\N}] \datatable;
            \addplot [name path=lower,draw=none] table[x index = {0},y expr=\thisrowno{1}-\thisrowno{2},skip coords between index={75}{\N}] \datatable;
            \addplot [fill=blue!30] fill between[of=upper and lower];
            \addplot [name path=upper,draw=none] table[x index = {0},y expr=\thisrowno{1}+\thisrowno{2},skip coords between index={75}{\N}] \datatabletwo;
            \addplot [name path=lower,draw=none] table[x index = {0},y expr=\thisrowno{1}-\thisrowno{2},skip coords between index={75}{\N}] \datatabletwo;
            \addplot [fill=red!30] fill between[of=upper and lower];
        \nextgroupplot[ylabel={Sequence no}]
            \addplot[mark=none, blue] table[x index = {0},y index = {3},skip coords between index={75}{\N}] \datatable;
            \addplot[mark=none, dashed, red] table[x index = {0},y index = {3},skip coords between index={75}{\N}] \datatabletwo;
            \addplot [name path=upper,draw=none] table[x index = {0},y expr=\thisrowno{3}+\thisrowno{4},skip coords between index={75}{\N}] \datatable;
            \addplot [name path=lower,draw=none] table[x index = {0},y expr=\thisrowno{3}-\thisrowno{4},skip coords between index={75}{\N}] \datatable;
            \addplot [fill=blue!20] fill between[of=upper and lower];
            \addplot [name path=upper,draw=none] table[x index = {0},y expr=\thisrowno{3}+\thisrowno{4},skip coords between index={75}{\N}] \datatabletwo;
            \addplot [name path=lower,draw=none] table[x index = {0},y expr=\thisrowno{3}-\thisrowno{4},skip coords between index={75}{\N}] \datatabletwo;
            \addplot [fill=red!20] fill between[of=upper and lower];
        \nextgroupplot[ylabel={Queue size}, xlabel={sim-time (s)}, scaled y ticks=base 10:-2,]
            \addplot[mark=none, blue] table[x index = {0},y index = {5},skip coords between index={75}{\N}] \datatable;
            \addplot[mark=none, dashed, red] table[x index = {0},y index = {5},skip coords between index={75}{\N}] \datatabletwo;
            \addplot [name path=upper,draw=none] table[x index = {0},y expr=\thisrowno{5}+\thisrowno{6},skip coords between index={75}{\N}] \datatable;
            \addplot [name path=lower,draw=none] table[x index = {0},y expr=\thisrowno{5}-\thisrowno{6},skip coords between index={75}{\N}] \datatable;
            \addplot [fill=blue!20] fill between[of=upper and lower];
            \addplot [name path=upper,draw=none] table[x index = {0},y expr=\thisrowno{5}+\thisrowno{6},skip coords between index={75}{\N}] \datatabletwo;
            \addplot [name path=lower,draw=none] table[x index = {0},y expr=\thisrowno{5}-\thisrowno{6},skip coords between index={75}{\N}] \datatabletwo;
            \addplot [fill=red!20] fill between[of=upper and lower];
    \end{groupplot}
\end{tikzpicture}}}
  \caption{Case 1 and 2: step=2ms, delay=2ms, thread=$t_{\mathit{off}}$/$t_{\mathit{on}}$ }
  \label{fig:log_f2_s2_z2_orig}
\end{figure}
\fi

\ifdefined\PAPER
In this paper we will present only the results from cases $3$ and $4$ (Table~\ref{tab:exp1}) due to space constraints, and refer the reader to the technical report~\cite{frasheri2021rmqfmu} that contains the results over all cases. The results not included in this paper are coherent with the ones included.
In the included scenarios, the simulation step and the simulation delay are set each to $100$ms, in order to mimic a co-simulation environment that imposes a step size of $100$ms and has a simulation delay also of $100$ms, caused e.g.\ by other FMUs or monitors. 
The results of these two tests are shown in Figure~\ref{fig:log_f2_s100_z100_orig}.
The number of messages to process in each step is $50$, given the input frequency of $2$ms. 
Additionally, given the simulation delay equal to $100$ms, for the threaded configuration all $50$ messages will be available in the RMQFMU\textsubscript{1} queue when the step function is called. The artificial simulation delay of $100$ms covers enough time for the separate consumer thread to have consumed approximately $50$ messages. While for the non-threaded configuration, the step function itself needs to consume approximately $50$ messages off the socket interface inside a single step. 
From Figure~\ref{fig:log_f2_s100_z100_orig} left-hand side graph it can be observed that for this test, the threaded configuration (red) shows an improvement in average step duration of approximately $10$ms compared to the unthreaded configuration (blue).
This difference corresponds to the overhead of consuming a single message from the socket interface via the rabbitmq client library, parsing the message, and finally adding it to the internal RMQFMU\textsubscript{1} queue. 
As approximately $50$ messages need to be consumed in a single step, this accounts to an overhead of approximately $200\mu$s per message ($10$ms $/$ $50$ = $200\mu$s). 
The right-hand side graph in Figure~\ref{fig:log_f2_s100_z100_orig} shows the internal FMU queue size at the exit of each simulation time step. The increase of queue size in the threaded configuration (red) is mostly an effect of gaps occurring in the input data. 
Large gaps in the input will cause the RMQFMU\textsubscript{1} to stay at its current output for a predefined \textit{maxage} time period ideally covering the input gap. 
This effect can also be observed in the left-hand side graph, when the step duration occasionally lowers to around $0$ when the RMQFMU\textsubscript{1} stays at its current output to cover a gap. 
For this period, data will still be consumed by the separate RMQFMU\textsubscript{1} consumer thread and added to the internal queue. 
The RMQFMU\textsubscript{1} implementation must therefore include guards to respect internal queue size limitations. 
The internal RMQFMU\textsubscript{1} queue size in the unthreaded configuration (blue) is always $0$ at simulation step exit, since that configuration consumes only a single message off the socket interface per step. 
In this configuration, a queue size build up will occur in the socket layer rather than the FMU layer.
%\casper{200 us - magic number??}
%\gita{Fixed.}
%\casper{No mention of the QUeue size, why include it?}
%\gita{ctoff to toff and explain what it stands for, turn ms to ms in the graph. }
%\casper{why? elaborate more on this point, might be difficult for the reader to get it.}
%\gita{say that the graphs that are not included, are coherent with what we have here, w/o going to too many details.}
%\henrik{Fixed. Did a rewrite of the above part regarding both the step duration and the queue size. Hopefully, addresses all comments. Please review.}
\fi

\ifdefined\TECHREP
First we consider the results of case 1 and case 2 from Table~\ref{tab:exp1} (Figure~\ref{fig:log_f2_s2_z2_orig}). 
These tests are carried out with a simulation step matching the rate of incoming data at $2$ms and with a minimal simulation delay of also $2$ms. 
The running tests collect profiling information from the RMQFMU\textsubscript{1} consisting for each simulation step of: the \textit{step duration}, which is the system execution time for the RMQFMU\textsubscript{1} \texttt{doStep} function, the \textit{sequence number} output, which is the message sequence number output by the RMQFMU\textsubscript{1}, and the \textit{queue size}, which is the size of the internal incoming message queue in the RMQFMU\textsubscript{1}. 
The tests are run $5$ times each to provide a minimal set of upper and lower bounds, and each test in this case runs for $1.5$sec simulation time.
%\henrik{@Gita: We should scale Fig.2 to cover only 1.5sec as it used to be. This to match the text here and to better show the initial 300ms behaviour, that we describe. This is even more important when reading the later paragrphs on this page, that relates gaps from Fig.4. I think also Fig.4 should be scaled to its original size, which I believe was also covering 1.5sec.}
%\gita{you mean the simulation runs for 1.5 sec right?}

A few things can be observed from the figure. 
First, it can be seen from the step duration graph (top, blue line), that the average step duration appears to be around $500$-$600$us with some drops to around $200$us in some periods. 
So, in general the step duration is lower than the $2$ms frequency of the input data, meaning that the RMQFMU\textsubscript{1} is capable of keeping up with the input data rate in this configuration.

The periods where the step duration drops to around $200$us are caused by two different reasons. 
The first period from simulation time $0$ to time $0.3$ is caused by the max age setting of $300$ms for this test. 
This max age means that the initial output value of the RMQFMU\textsubscript{1} is considered valid for the first $300$ms and hence no consumption of new RabbitMQ messages are needed for this period and as a result the step duration is low. From the middle graph (blue line) observe, that the unique message sequence number output by the RMQFMU\textsubscript{1} stays at the initial value for the first $300$ms. 
%Once the max age has passed, the RMQFMU\textsubscript{1} needs to consume data from the RabbitMQ interface in each $2$ms step. 
%This amounts to a cost of around $200-300$us per message to consume the RabbitMQ message from the client socket interface. 
%Leaving the total step duration at around $500-600$us.
%\casper{Not understood}
%\henrik{I have left out what was the final two sentences from above. That was not there in the original text, and I think the point about the per message cost of consuming from the socket interface becomes better explained anyway on p.12. Please review.}
The second cause for the intermittent step duration drops as seen in the top graph (e.g. starting between time step $0.9$ and $1.0$) is due to time gaps in the input data. There are multiple occasions of consecutive input data messages being separated by much more than $2$ms in the test input data csv file. 
%Observe e.g. a larger gap of +$40$ms around time step $0.9$. 
%This gap is the reason for the step duration dropping in Figure~\ref{fig:log_f2_s2_z2_orig} (top graph, blue line?) at this time. 
For the simulation steps happening during these gaps, no new input is provided and the max age causes the RMQFMU\textsubscript{1} to stay at its current output and hence the step duration is low. Finally, note from the bottom graph (blue line), that the RMQFMU\textsubscript{1} input queue size is constantly at $0$ after a simulation step has finished, since each simulation step consumes all (in this case $1$) messages from the RMQFMU\textsubscript{1} queue. However, the rabbitmq client library socket layer incoming queue may still increase. 

\ifdefined\TECHREP
\begin{figure}[tbh!]
  \centerline{{\begin{tikzpicture}
    \pgfplotstableread[col sep=comma]{raw_data/rbmqv1/urdata/ctoff_s100_z100.csv}\datatable
    \pgfplotstableread[col sep=comma]{raw_data/rbmqv1/urdata/cton_s100_z100.csv}\datatabletwo
    
    \pgfplotstablegetrowsof{\datatable}
    \pgfmathsetmacro{\N}{\pgfplotsretval} 
    \begin{groupplot}[group style={
                    group name=myplot,
                    group size= 1 by 3,vertical sep=1cm},height=4cm,width=8cm,
                    legend style={at={(0.5,-3.3)},anchor=north,legend columns=-1},
                    xmin=-0.1,
                    xmax=20.1,
                    minor x tick num = 4,
                    %minor y tick num = 4,
                    %xminorgrids=true,
                    %yminorgrids=true,
                    %xmajorgrids=true,
                    %ymajorgrids=true,
                    scaled y ticks=base 10:-3,
                    ]
        \nextgroupplot[ylabel={Step duration ($\mu s$)},ymax=20000]
            \addplot[mark=none, blue] table[x index = {0},y index = {1}] \datatable;
            \addplot[mark=none, dashed, red] table[x index = {0},y index = {1}] \datatabletwo;
            \addlegendentry{{$t_{\mathit{off}}$}}
            \addlegendentry{{$t_{\mathit{on}}$}}
            \addplot [name path=upper,draw=none] table[x index = {0},y expr=\thisrowno{1}+\thisrowno{2}] \datatable;
            \addplot [name path=lower,draw=none] table[x index = {0},y expr=\thisrowno{1}-\thisrowno{2}] \datatable;
            \addplot [fill=blue!20] fill between[of=upper and lower];
            \addplot [name path=upper,draw=none] table[x index = {0},y expr=\thisrowno{1}+\thisrowno{2}] \datatabletwo;
            \addplot [name path=lower,draw=none] table[x index = {0},y expr=\thisrowno{1}-\thisrowno{2}] \datatabletwo;
            \addplot [fill=red!20] fill between[of=upper and lower];

        \nextgroupplot[ylabel={Sequence no}]
            \addplot[mark=none, blue] table[x index = {0},y index = {3}] \datatable;
            \addplot[mark=none, dashed, red] table[x index = {0},y index = {3}] \datatabletwo;
            \addplot [name path=upper,draw=none] table[x index = {0},y expr=\thisrowno{3}+\thisrowno{4}] \datatable;
            \addplot [name path=lower,draw=none] table[x index = {0},y expr=\thisrowno{3}-\thisrowno{4}] \datatable;
            \addplot [fill=blue!20] fill between[of=upper and lower];
            \addplot [name path=upper,draw=none] table[x index = {0},y expr=\thisrowno{3}+\thisrowno{4}] \datatabletwo;
            \addplot [name path=lower,draw=none] table[x index = {0},y expr=\thisrowno{3}-\thisrowno{4}] \datatabletwo;
            \addplot [fill=red!20] fill between[of=upper and lower];

        \nextgroupplot[ylabel={Queue size}, xlabel={sim-time (s)}, ymax=800,scaled y ticks=base 10:-2,]
            \addplot[mark=none, blue] table[x index = {0},y index = {5}] \datatable;
            \addplot[mark=none, dashed, red] table[x index = {0},y index = {5}] \datatabletwo; 
            \addplot [name path=upper,draw=none] table[x index = {0},y expr=\thisrowno{5}+\thisrowno{6}] \datatable;
            \addplot [name path=lower,draw=none] table[x index = {0},y expr=\thisrowno{5}-\thisrowno{6}] \datatable;
            \addplot [fill=blue!20] fill between[of=upper and lower];
            \addplot [name path=upper,draw=none] table[x index = {0},y expr=\thisrowno{5}+\thisrowno{6}] \datatabletwo;
            \addplot [name path=lower,draw=none] table[x index = {0},y expr=\thisrowno{5}-\thisrowno{6}] \datatabletwo;
            \addplot [fill=red!20] fill between[of=upper and lower];
    \end{groupplot}
\end{tikzpicture}}}
  \caption{Cases 3 and 4: step=100ms, delay=100ms, thread=$t_{\mathit{off}}$/$t_{\mathit{on}}$ }
  \label{fig:log_f2_s100_z100_orig}
\end{figure}
\fi
\ifdefined\PAPER
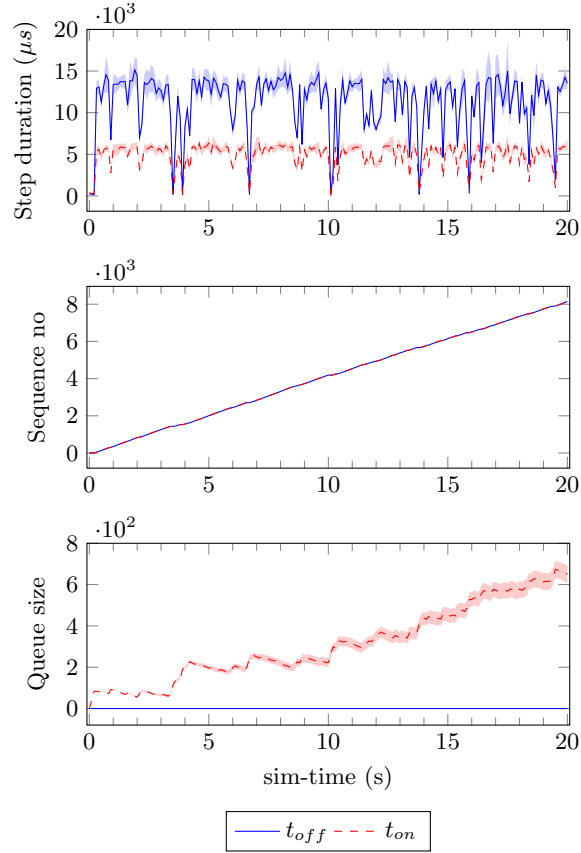
\begin{figure}[t]
  \centerline{\resizebox{.65\textwidth}{!}{\begin{tikzpicture}
    \pgfplotstableread[col sep=comma]{raw_data/rbmqv1/urdata/ctoff_s100_z100.csv}\datatable
    \pgfplotstableread[col sep=comma]{raw_data/rbmqv1/urdata/cton_s100_z100.csv}\datatabletwo
    
    \pgfplotstablegetrowsof{\datatable}
    \pgfmathsetmacro{\N}{\pgfplotsretval} 
    \begin{groupplot}[group style={
                    group name=myplot,
                    group size= 1 by 3,vertical sep=1cm},height=4cm,width=8cm,
                    legend style={at={(0.5,-3.3)},anchor=north,legend columns=-1},
                    xmin=-0.1,
                    xmax=20.1,
                    minor x tick num = 4,
                    %minor y tick num = 4,
                    %xminorgrids=true,
                    %yminorgrids=true,
                    %xmajorgrids=true,
                    %ymajorgrids=true,
                    scaled y ticks=base 10:-3,
                    ]
        \nextgroupplot[ylabel={Step duration ($\mu s$)},ymax=20000]
            \addplot[mark=none, blue] table[x index = {0},y index = {1}] \datatable;
            \addplot[mark=none, dashed, red] table[x index = {0},y index = {1}] \datatabletwo;
            \addlegendentry{{$t_{\mathit{off}}$}}
            \addlegendentry{{$t_{\mathit{on}}$}}
            \addplot [name path=upper,draw=none] table[x index = {0},y expr=\thisrowno{1}+\thisrowno{2}] \datatable;
            \addplot [name path=lower,draw=none] table[x index = {0},y expr=\thisrowno{1}-\thisrowno{2}] \datatable;
            \addplot [fill=blue!20] fill between[of=upper and lower];
            \addplot [name path=upper,draw=none] table[x index = {0},y expr=\thisrowno{1}+\thisrowno{2}] \datatabletwo;
            \addplot [name path=lower,draw=none] table[x index = {0},y expr=\thisrowno{1}-\thisrowno{2}] \datatabletwo;
            \addplot [fill=red!20] fill between[of=upper and lower];

        \nextgroupplot[ylabel={Sequence no}]
            \addplot[mark=none, blue] table[x index = {0},y index = {3}] \datatable;
            \addplot[mark=none, dashed, red] table[x index = {0},y index = {3}] \datatabletwo;
            \addplot [name path=upper,draw=none] table[x index = {0},y expr=\thisrowno{3}+\thisrowno{4}] \datatable;
            \addplot [name path=lower,draw=none] table[x index = {0},y expr=\thisrowno{3}-\thisrowno{4}] \datatable;
            \addplot [fill=blue!20] fill between[of=upper and lower];
            \addplot [name path=upper,draw=none] table[x index = {0},y expr=\thisrowno{3}+\thisrowno{4}] \datatabletwo;
            \addplot [name path=lower,draw=none] table[x index = {0},y expr=\thisrowno{3}-\thisrowno{4}] \datatabletwo;
            \addplot [fill=red!20] fill between[of=upper and lower];

        \nextgroupplot[ylabel={Queue size}, xlabel={sim-time (s)}, ymax=800,scaled y ticks=base 10:-2,]
            \addplot[mark=none, blue] table[x index = {0},y index = {5}] \datatable;
            \addplot[mark=none, dashed, red] table[x index = {0},y index = {5}] \datatabletwo; 
            \addplot [name path=upper,draw=none] table[x index = {0},y expr=\thisrowno{5}+\thisrowno{6}] \datatable;
            \addplot [name path=lower,draw=none] table[x index = {0},y expr=\thisrowno{5}-\thisrowno{6}] \datatable;
            \addplot [fill=blue!20] fill between[of=upper and lower];
            \addplot [name path=upper,draw=none] table[x index = {0},y expr=\thisrowno{5}+\thisrowno{6}] \datatabletwo;
            \addplot [name path=lower,draw=none] table[x index = {0},y expr=\thisrowno{5}-\thisrowno{6}] \datatabletwo;
            \addplot [fill=red!20] fill between[of=upper and lower];
    \end{groupplot}
\end{tikzpicture}}}
  \caption{Cases 3 and 4: step=100ms, delay=100ms, thread=$t_{\mathit{off}}$/$t_{\mathit{on}}$ }
  \label{fig:log_f2_s100_z100_orig}
\end{figure}
\fi

%\begin{figure}
% \centerline{{\input{Figures/rbmqv1/urdata/timegaps}}}
%
%  \caption{Time gaps in input data}
% \label{fig:time_gaps}
%\end{figure}

Now consider the same test but with the FMU configured to use the threaded setting. The results of this case are shown in Figure~\ref{fig:log_f2_s2_z2_orig}. Notice that the step duration (top, red dashed line) is now lower compared to the non-threaded case. From the graph, the step duration appears to be around $300-400$us. 
Compared to the non-threaded case with $500-600$us step durations, this is approximately $200-300$us lower. 
Which corresponds to the additional cost for consuming a single message from the RabbitMQ socket interface and parsing the JSON message format for the 117 message value references and placing the parsed message content in the internal FMU queue. 
In the threaded configuration, this message consumption executes in the separate consumer thread partly while the main thread delays for the configured $2$ms. 
\fi

\ifdefined\TECHREP
%\begin{figure}
%  \centerline{\includegraphics[scale=0.60]{Figures/log_cton_f2_s2_z2_orig.pdf}}
%  \caption{Case 2: step=2ms, delay=2ms, thread=on}
%  \label{fig:log_cton_f2_s2_z2_orig}
%\end{figure}

The same tests as above are also considered for cases 3 and 4, where the simulation step and the simulation delay is increased from $2$ms to $100$ms. 
This is done to mimic a co-simulation environment that imposes a step size of $100$ms and has a simulation delay also of $100$ms, caused e.g.\ by other FMUs or monitors. 
The results of these two tests are shown in Figure~\ref{fig:log_f2_s100_z100_orig}.

The results are equivalent to the results for cases 1 and 2. But as the simulation step size has increased to $100$ms, the number of messages to process in each step has now increased from $1$ to $50$ since the input frequency is still $2$ms. 
And since the simulation delay has also increased to $100$ms, it means that for the threaded configuration basically all $50$ messages will be available in the FMU queue when the step function is called. 
While for the non-threaded configuration, the step function itself needs to consume approximately $50$ messages of the socket interface inside a single step. 
As approximately $50$ messages need to be consumed in a single step, this accounts to an overhead of approximately $200\mu$s per message ($10$ms $/$ $50$ = $200\mu$s). 
And this corresponds to the observed difference in average step duration between the blue and red lines in Figure~\ref{fig:log_f2_s100_z100_orig}. So, for this configuration, the threaded configuration has a performance improvement of around $10$ms per step.
\fi

\ifdefined\TECHREP
Also notice from Figure~\ref{fig:log_f2_s100_z100_orig}, that the queue size continually increases at the time points where the input data has time gaps as the FMU consumes no new inputs. The increase of queue size in the threaded configuration (red) is mostly an effect of gaps occurring in the input data. Large gaps in the input will cause the FMU to stay at its current output for a predefined \textit{maxage} time period ideally covering the input gap. 
For this case-study, data is published with a fixed interval of $2$ms between consecutive messages, even though the message timestamps may be having larger gaps. I.e.\ this is an example of replaying historical data with a fixed message rate, as opposed to sending live data. Hence, for the gap periods, data will still be consumed by the separate FMU consumer thread and added to the internal queue. 
The FMU implementation must therefore include guards to respect internal queue size limitations.
As discussed, the basic condition for a potential performance improvement using the threaded configuration will be present if the simulation step size is smaller than the simulation delay. 
If this condition is present, the number of messages to process in a simulation step may be already present in the FMU queue at the start of the step. 
And the total improvement gain is linear in the number of messages present in the queue. 
In our tests with a factor of around $200$us per message. 
The number of messages in the queue to be processed at each simulation step is determined by the relation between the input data frequency and the simulation step size.

%\begin{figure}
%  \centerline{\includegraphics[scale=0.60]{Figures/log_cton_f2_s100_z100_orig.pdf}}
%  \caption{Case 4a: step=100ms, delay=100ms, thread=on}
%  \label{fig:log_cton_f2_s100_z100_orig}
%\end{figure}
\begin{figure}[tbh!]
  \centerline{{\begin{tikzpicture}
    \pgfplotstableread[col sep=comma]{raw_data/rbmqv1/urdata/cton_s100_z113.csv}\datatable
    \pgfplotstableread[col sep=comma]{raw_data/rbmqv1/urdata/cton_s100_z120.csv}\datatabletwo
    \pgfplotstableread[col sep=comma]{raw_data/rbmqv1/urdata/cton_s100_z100_clean.csv}\datatablethree
    \pgfplotstablegetrowsof{\datatable}
    \pgfmathsetmacro{\N}{\pgfplotsretval} 
    \begin{groupplot}[group style={
                    group name=myplot,
                    group size= 1 by 2,vertical sep=1cm},height=4cm,width=8cm,
                    legend style={at={(0.5,-2.)},anchor=north,legend columns=-1},
                    xmin=-0.1,
                    xmax=20.1,
                    minor x tick num = 4,
                    %minor y tick num = 4,
                    %xminorgrids=true,
                    %yminorgrids=true,
                    %xmajorgrids=true,
                    %ymajorgrids=true,
                    scaled y ticks=base 10:-3,
                    ]
        \nextgroupplot[ylabel={Step duration ($\mu s$)},ymax=29000]

            \addplot[mark=none, dash dot, green] table[x index = {0},y index = {1}] \datatablethree;
            \addplot[mark=none, blue] table[x index = {0},y index = {1}] \datatable;
            \addplot[mark=none, dashed, red] table[x index = {0},y index = {1}] \datatabletwo;
            
            \addlegendentry{{$d = 100$}}
            \addlegendentry{{$d = 113$}}
            \addlegendentry{{$d = 120$}}

            \addplot [name path=upper,draw=none] table[x index = {0},y expr=\thisrowno{1}+\thisrowno{2}] \datatablethree;
            \addplot [name path=lower,draw=none] table[x index = {0},y expr=\thisrowno{1}-\thisrowno{2}] \datatablethree;
            \addplot [fill=green!20] fill between[of=upper and lower];
            
            \addplot [name path=upper,draw=none] table[x index = {0},y expr=\thisrowno{1}+\thisrowno{2}] \datatable;
            \addplot [name path=lower,draw=none] table[x index = {0},y expr=\thisrowno{1}-\thisrowno{2}] \datatable;
            \addplot [fill=blue!20] fill between[of=upper and lower];

            \addplot [name path=upper,draw=none] table[x index = {0},y expr=\thisrowno{1}+\thisrowno{2}] \datatabletwo;
            \addplot [name path=lower,draw=none] table[x index = {0},y expr=\thisrowno{1}-\thisrowno{2}] \datatabletwo;
            \addplot [fill=red!20] fill between[of=upper and lower];
            
        %\nextgroupplot[ylabel={Sequence no}]
        
         %   \addplot[mark=none, green, very thick] table[x index = {0},y index = {3}] \datatablethree;
          %  \addplot[mark=none, dash dot, blue] table[x index = {0},y index = {3}] \datatable;
         %   \addplot[mark=none, dashed, red] table[x index = {0},y index = {3}] \datatabletwo;
            
         %   \addplot [name path=upper,draw=none] table[x index = {0},y expr=\thisrowno{3}+\thisrowno{4}] \datatablethree;
         %   \addplot [name path=lower,draw=none] table[x index = {0},y expr=\thisrowno{3}-\thisrowno{4}] \datatablethree;
         %   \addplot [fill=green!20] fill between[of=upper and lower];
            
         %   \addplot [name path=upper,draw=none] table[x index = {0},y expr=\thisrowno{3}+\thisrowno{4}] \datatable;
         %   \addplot [name path=lower,draw=none] table[x index = {0},y expr=\thisrowno{3}-\thisrowno{4}] \datatable;
         %   \addplot [fill=blue!20] fill between[of=upper and lower];

         %   \addplot [name path=upper,draw=none] table[x index = {0},y expr=\thisrowno{3}+\thisrowno{4}] \datatabletwo;
         %   \addplot [name path=lower,draw=none] table[x index = {0},y expr=\thisrowno{3}-\thisrowno{4}] \datatabletwo;
         %   \addplot [fill=red!20] fill between[of=upper and lower];
            
        \nextgroupplot[ylabel={Queue size}, xlabel={sim-time (s)}, ymax=800,scaled y ticks=base 10:-2,]
        
            \addplot[mark=none, dash dot, green] table[x index = {0},y index = {5}] \datatablethree;
            \addplot[mark=none, blue] table[x index = {0},y index = {5}] \datatable;
            \addplot[mark=none, dashed, red] table[x index = {0},y index = {5}] \datatabletwo;
            
            \addplot [name path=upper,draw=none] table[x index = {0},y expr=\thisrowno{5}+\thisrowno{6}] \datatablethree;
            \addplot [name path=lower,draw=none] table[x index = {0},y expr=\thisrowno{5}-\thisrowno{6}] \datatablethree;
            \addplot [fill=green!20] fill between[of=upper and lower];
            
            \addplot [name path=upper,draw=none] table[x index = {0},y expr=\thisrowno{5}+\thisrowno{6}] \datatable;
            \addplot [name path=lower,draw=none] table[x index = {0},y expr=\thisrowno{5}-\thisrowno{6}] \datatable;
            \addplot [fill=blue!20] fill between[of=upper and lower];

            \addplot [name path=upper,draw=none] table[x index = {0},y expr=\thisrowno{5}+\thisrowno{6}] \datatabletwo;
            \addplot [name path=lower,draw=none] table[x index = {0},y expr=\thisrowno{5}-\thisrowno{6}] \datatabletwo;
            \addplot [fill=red!20] fill between[of=upper and lower];

    \end{groupplot}

\end{tikzpicture}}}
  \caption{Case 5 and 6: step=100ms, delay=113 and 120ms, thread=on, cleaned input}
  \label{fig:log_cton_f2_s100_z_clean}
\end{figure}
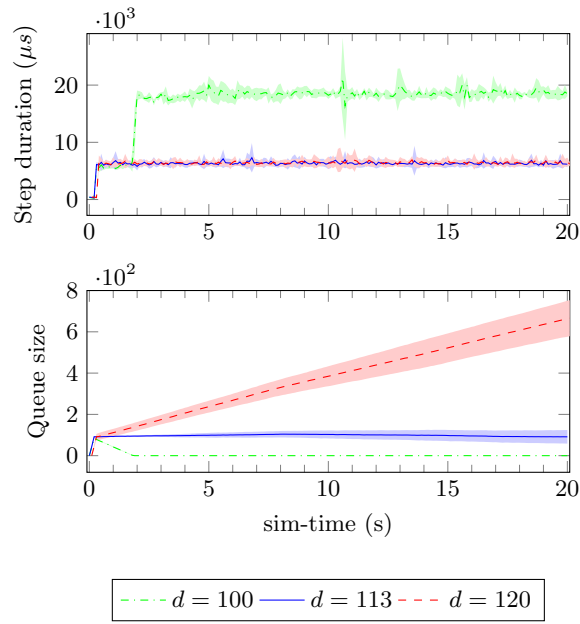

The optimal relation between simulation step size and delay is very much system dependent and will vary. 
To observe the effects of increased simulation delays, we provide results for three tests with a fixed step size of $100$ms and three increasing delays of $100$ms, $113$ms, and $120$ms. Furtermore, as described earlier, the time gaps in the input data will reflect in the test results as time points where the FMU will not consume inputs and the output will be unchanged. 
And since the test environment still replays the delayed messages with a fixed interval (here of $2$ms), the input queue size will increase in these points. 
As a result, the FMU may have improved step duration caused by the fact of having time gaps in the input while still replaying it with a fixed interval. 
To remove the effect of this issue, these final tests are using a ``cleaned'' data set, meaning that there is a fixed timestamp interval of exactly $2$ms between all messages. I.e.\ the data has been "cleaned" from gaps.
The results are shown in Figures~\ref{fig:log_cton_f2_s100_z_clean}. 
%\casper{These figures should not be referenced without the following paragraph explained beforehand}
%\henrik{Fixed. Reference moved till later. Please review.}
\fi
\ifdefined\TECHREP
%\lukas{queue size in purple (fig 10, 11, 12 -- previously in orange}
%\henrik{Not sure what is meant here? In Fig.5 the queue size is in three diff colors. Please clarify.}
%\begin{figure}
%  \centerline{\includegraphics[scale=0.60]{Figures/log_cton_f2_s100_z100_clean.pdf}}
%  \caption{Case 4b: step=100ms, delay=100ms, thread=on, cleaned input}
%  \label{fig:log_cton_f2_s100_z100_clean}
%\end{figure}
%\lukas{what means `cleaned input'?}
%\henrik{Fixed. Should be clear now from the final lines of above paragraph.}

%\begin{figure}
%  \centerline{\includegraphics[scale=0.60]{Figures/log_cton_f2_s100_z113_clean.pdf}}
%  \caption{Case 5: step=100ms, delay=113ms, thread=on, cleaned input}
%  \label{fig:log_cton_f2_s100_z113_clean}
%\end{figure}

The test results with a simulation delay of $100$ms is shown in Figure~\ref{fig:log_cton_f2_s100_z_clean} (green line). 
From this test it can be observed, that the $100$ms delay is actually too close to the step size to provide enough time for the input queue too constantly be filled with enough messages to process in the simulation step. 
The results show, that the input queue increases initially, but within the first $2$ seconds, the simulation has stepped fast enough for the FMU to gradually empty its queue. 
From that point onwards, the FMU will not have enough messages present in its queue to cover the entire simulation step. 
Hence, it needs to block for I/O to obtain new messages, and thus the step duration increases.
Increasing the tested simulation delay to $113$ms provides just enough additional delay time for the input queue to have enough data for the tested time range of $20$sec (Figure~\ref{fig:log_cton_f2_s100_z_clean} blue line). 
Notice that the step duration is constantly lower compared to the previous test. 
Finally, increasing the tested simulation delay to $120$ms provides no additional improvement in the step duration, but rather has the effect of having the FMU input queue increasing continually (Figure~\ref{fig:log_cton_f2_s100_z_clean} dashed red line). 

%HE: XXX to be removed
\fi

\subsubsection{Case-study 2}
\ifdefined\TECHREP
Results for cases $1-4$ are displayed in Figure~\ref{fig:fig1}.
It is possible to observe that for a low value of the maxage ($200$ms), there is no impact from changing the value of the lookahead.
In contrast, for higher values of the maxage ($2$s), a higher value of the lookahead allows to jump to the latest messages valid for the time-step that we want to reach  -- those steps for which the current values are not valid with respect to the maxage, thus a new value is needed.

Whereas, a low value of the lookahead will result in older messages being outputted by the RMQFMU\textsubscript{1}.
The message that is outputted at a particular time-step depends on what is available in the queue at that particular time. 
We can observe in Figure~\ref{fig:fig2}, with a lookahead equal to $50$ and maxage fixed to $2000$ms, over 5 independent runs, that the number of skipped messages is not the exactly the same from run to run..
One factor that impacts the state of the queue is the speed of the execution of a simulation step. 
In Figure~\ref{fig:fig2}, one can observe the wall-clock time for two of the simulation runs. 
The first run is executed faster than real-time, at least until until time step $2.0$, whereas the second is executed slower than real-time. 
This depends on the underlying execution environment, as well as the load on said environment.
As such, a specific lookahead will not necessarily output the same sequence numbers at each time step over different independent runs. 
%\gita{it might be that in some cases the co-simulation reaches the 2 second mark faster than real-time, which means that it will have less than 20 msgs to consider. Note that data from the python script is sent at 100 ms intervals.}
%\gita{This seems to be the case. paragraph needs to be updated to reflect this.}

Results for cases $5-8$ can be observed in Figure~\ref{fig:fig3}, whereas for cases $9-12$ in Figure~\ref{fig:fig4}, consistent with what is observed in Figure~\ref{fig:fig1}.
\fi
\ifdefined\PAPER
Similarly to Case-study 1, we will present only the results from cases $9-12$ (Table~\ref{tab:exp2}) due to space constraints, and refer the reader to the technical report~\cite{frasheri2021rmqfmu} that contains the results over all cases. 
Note that the results not included remain coherent with the ones presented in the current paper.
The results for cases $9-12$ are displayed in Figure~\ref{fig:fig4}.
\fi
For a \textit{maxage} equal to $400$ms, we can see that there is not much an effect of the value set for the \textit{lookahead}.
This is due to the fact that with lower \textit{maxage}, the data becomes invalid sooner than for higher \textit{maxage}, thus triggering the RMQFMU\textsubscript{1} to fetch newer data from the incoming queue.
%\gita{explain why?}
Nevertheless, for a \textit{maxage} equal to $2$s, it is possible to note a difference for the two \textit{lookahead} values, where for $la=5$, we can see a bigger jump in the sequences of data for specific time-steps, when a new value is needed.
Additionally, sequence numbers remain fixed for a consecutive number of time-steps, whereas for $la=2$ more in-between value are outputted, as expected.
Nevertheless, in the latter, there is more delay in the output values.
\ifdefined\TECHREP
\begin{figure}[tbh!]
    \centering
    \begin{tikzpicture}
    \pgfplotstableread[col sep=comma]{raw_data/rbmqv1/gazebo/outputs_200_1.csv}\datatable
    \pgfplotstableread[col sep=comma]{raw_data/rbmqv1/gazebo/outputs_200_50.csv}\datatabletwo
    \pgfplotstableread[col sep=comma]{raw_data/rbmqv1/gazebo/outputs_2000_1.csv}\datatablethree
    \pgfplotstableread[col sep=comma]{raw_data/rbmqv1/gazebo/outputs_2000_50_run1.csv}\datatablefour
    \begin{groupplot}[group style={
                    group name=myplot,
                    group size= 2 by 1,vertical sep=1.cm},
                    height=5cm, width=6cm,
                    legend style={at={(1.08,-0.3)},anchor=north,legend columns=-1},
                    minor x tick num = 2,
                    %minor y tick num = 4,
                    %xminorgrids=true,
                    %yminorgrids=true,
                    %xmajorgrids=true,
                    %ymajorgrids=true,
                    ]
        \nextgroupplot[title={ma=$200$ms},ylabel={Sequence numbers}, xlabel={sim-time}]
            \addplot[only marks, mark=x, blue] table [x index = {0}, y index = {5}] \datatable;
            \addplot[only marks, mark=+,red] table [x index = {0}, y index = {5}] \datatabletwo;
        \addlegendentry{{$la=1$}}
        \addlegendentry{{$la=50$}}
        \nextgroupplot[title={ma=$2000$ms}, xlabel={sim-time}]
            \addplot[only marks, mark=x, blue] table [x index = {0}, y index = {5}] \datatablethree;
            \addplot[only marks, mark=+,red] table [x index = {0}, y index = {5}] \datatablefour;
            
    \end{groupplot}
    %\node (title) at ($(myplot c1r1.center)!0.5!(myplot c2r1.center)+(0,4cm)$) {Sequence numbers over time for $f_{data}=100$ms};
\end{tikzpicture}
    
    \caption{Messages outputted by the RMQFMU\textsubscript{1} at every step, with varying values of the \textit{maxage} and the \textit{lookahead}, for $f_{data}=100$ms.}
    \label{fig:fig1}
\end{figure}

\begin{figure}[tbh!]
    \centering
   \begin{tikzpicture}
    \pgfplotstableread[col sep=comma]{raw_data/rbmqv1/gazebo/outputs_ma200_la50_f2ms.csv}\datatable
    \pgfplotstableread[col sep=comma]{raw_data/rbmqv1/gazebo/outputs_ma200_la200_f2ms.csv}\datatabletwo
    \pgfplotstableread[col sep=comma]{raw_data/rbmqv1/gazebo/outputs_ma2000_la50_f2ms.csv}\datatablethree
    \pgfplotstableread[col sep=comma]{raw_data/rbmqv1/gazebo/outputs_ma2000_la200_f2ms.csv}\datatablefour

    \begin{groupplot}[group style={
                    group name=myplot,
                    group size= 2 by 1,vertical sep=1.5cm},
                    height=5cm, width=6cm,
                    legend style={at={(1.08,-0.3)},anchor=north,legend columns=-1},
                    minor x tick num = 2,
                    %minor y tick num = 4,
                    %xminorgrids=true,
                    %yminorgrids=true,
                    %xmajorgrids=true,
                    %ymajorgrids=true,
                    scaled y ticks=base 10:-3
                    ]
        \nextgroupplot[title={ma=$200$ms},ylabel={Sequence numbers}, xlabel={sim-time}]
            \addplot[only marks, mark=x, blue] table [x index = {0}, y index = {5}] \datatable;
            \addplot[only marks, mark=+,red] table [x index = {0}, y index = {5}] \datatabletwo;
        \addlegendentry{{$la=50$}}
        \addlegendentry{{$la=200$}}
        \nextgroupplot[title={ma=$2000$ms}, xlabel={sim-time}]
            \addplot[only marks, mark=x, blue] table [x index = {0}, y index = {5}] \datatablethree;
            \addplot[only marks, mark=+,red] table [x index = {0}, y index = {5}] \datatablefour;

    \end{groupplot}
    %\node (title) at ($(myplot c1r1.center)!0.5!(myplot c2r1.center)+(0,4cm)$) {Sequence numbers over time for $f_{data}=2$ms};
\end{tikzpicture}
    
    \caption{Messages outputted by the RMQFMU\textsubscript{1} at every step, with varying values of the \textit{maxage} and the \textit{lookahead}, for $f_{data}=2$ms.}
    \label{fig:fig3}
\end{figure}
\fi

\begin{figure}[tbh!]
    \centering
    \ifdefined\PAPER
    {\begin{tikzpicture}
    \pgfplotstableread[col sep=comma]{raw_data/rbmqv1/gazebo/outputs_ma400_la2_f200ms.csv}\datatable
    \pgfplotstableread[col sep=comma]{raw_data/rbmqv1/gazebo/outputs_ma400_la5_f200ms.csv}\datatabletwo
    \pgfplotstableread[col sep=comma]{raw_data/rbmqv1/gazebo/outputs_ma2000_la2_f200ms.csv}\datatablethree
    \pgfplotstableread[col sep=comma]{raw_data/rbmqv1/gazebo/outputs_ma2000_la5_f200ms.csv}\datatablefour

    \begin{groupplot}[group style={
                    group name=myplot,
                    group size= 2 by 1,horizontal sep=.8cm},height=3.5cm,width=4.5cm,
                    legend style={at={(2.7,1)},anchor=north,legend columns=1},
                    minor x tick num = 1,
                    %minor y tick num = 4,
                    %xminorgrids=true,
                    %yminorgrids=true,
                    %xmajorgrids=true,
                    %ymajorgrids=true,
                    ]
        \nextgroupplot[title={ma=$400$ms},ylabel={Sequence numbers}, xlabel={sim-time (s)}]
            \addplot[no marks, thick, blue] table [x index = {0}, y index = {5}] \datatable;
            \addplot[no marks, solid,red] table [x index = {0}, y index = {5}] \datatabletwo;
        \addlegendentry{{$la=2$}}
        \addlegendentry{{$la=5$}}
        \nextgroupplot[title={ma=$2000$ms},xlabel={sim-time (s)}]
            \addplot[no marks, thick, blue] table [x index = {0}, y index = {5}] \datatablethree;
            \addplot[no marks, solid,red] table [x index = {0}, y index = {5}] \datatablefour;

    \end{groupplot}
    %\node (title) at ($(myplot c1r1.center)!0.5!(myplot c2r1.center)+(0,01cm)$) {Sequence numbers over time for $f_{data}=200$ms};
\end{tikzpicture}}
    \fi
    \ifdefined\TECHREP
    {\begin{tikzpicture}
    \pgfplotstableread[col sep=comma]{raw_data/rbmqv1/gazebo/outputs_ma400_la2_f200ms.csv}\datatable
    \pgfplotstableread[col sep=comma]{raw_data/rbmqv1/gazebo/outputs_ma400_la5_f200ms.csv}\datatabletwo
    \pgfplotstableread[col sep=comma]{raw_data/rbmqv1/gazebo/outputs_ma2000_la2_f200ms.csv}\datatablethree
    \pgfplotstableread[col sep=comma]{raw_data/rbmqv1/gazebo/outputs_ma2000_la5_f200ms.csv}\datatablefour

    \begin{groupplot}[group style={
                    group name=myplot,
                    group size= 2 by 1,vertical sep=1.cm},
                    height=5cm, width=6cm,
                    legend style={at={(1.08,-0.3)},anchor=north,legend columns=-1},
                    minor x tick num = 1,
                    %minor y tick num = 4,
                    %xminorgrids=true,
                    %yminorgrids=true,
                    %xmajorgrids=true,
                    %ymajorgrids=true,
                    ]
        \nextgroupplot[title={ma=$400$ms},ylabel={Sequence numbers}, xlabel={sim-time (s)}]
            \addplot[only marks, mark=x, blue] table [x index = {0}, y index = {5}] \datatable;
            \addplot[only marks, mark=+,red] table [x index = {0}, y index = {5}] \datatabletwo;
        \addlegendentry{{$la=2$}}
        \addlegendentry{{$la=5$}}
        \nextgroupplot[title={ma=$2000$ms},xlabel={sim-time (s)}]
            \addplot[only marks, mark=x, blue] table [x index = {0}, y index = {5}] \datatablethree;
            \addplot[only marks, mark=+,red] table [x index = {0}, y index = {5}] \datatablefour;

    \end{groupplot}
    %\node (title) at ($(myplot c1r1.center)!0.5!(myplot c2r1.center)+(0,01cm)$) {Sequence numbers over time for $f_{data}=200$ms};
\end{tikzpicture}}
    \fi
    \caption{Messages outputted by the RMQFMU\textsubscript{1} at every step, with varying values of the \textit{maxage} and the \textit{lookahead}, for $f_{data}=200$ms.}
    \label{fig:fig4}
\end{figure}

The number of skipped messages is not the exactly the same from run to run, as observed in Figure~\ref{fig:fig2}, with a \textit{lookahead} equal to $50$ and \textit{maxage} fixed to $2000$ms, over 5 independent runs.
One factor that impacts the state of the queue is the speed of the execution of a simulation step. 
In Figure~\ref{fig:fig2}, one can observe the wall-clock time for two of the simulation runs. 
The first run is executed faster than real-time, at least until until time step $2.0$, whereas the second is executed slower than real-time. 
This depends on the underlying execution environment, as well as the load on said environment.
As such, a specific \textit{lookahead} will not necessarily output the same sequence numbers at each time step over different independent runs. 
\ifdefined\PAPER
This effect is also visible in Figure~\ref{fig:fig4}, where near the end, for both \textit{lookahead} values, the sequence numbers that are outputted are similar.
This means that, there were no newer messages in the queue at those particular times for \textit{lookahead}$=5$, as the simulation was running faster than real-time.
\fi
\ifdefined\TECHREP
This effect is also visible in Figures~\ref{fig:fig1} and~\ref{fig:fig4}, where for the last seconds, for both values of the \textit{lookahead}, the sequence numbers that are outputted are similar.
This means that, there were no newer messages in the queue at those particular times, as the simulation was running faster than real-time.
\fi

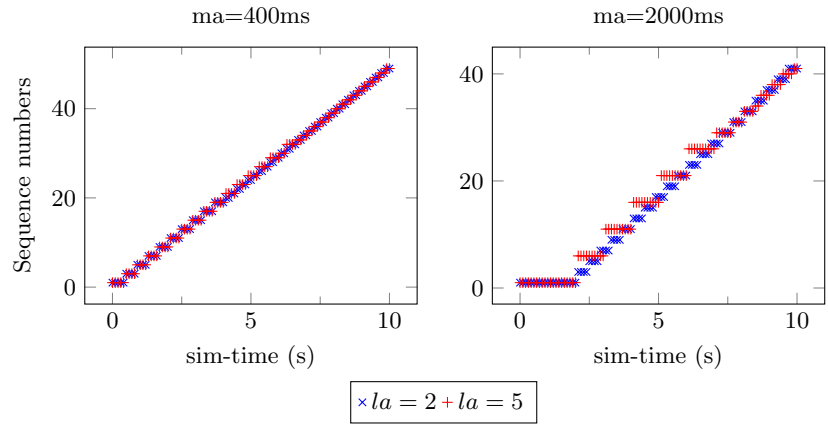
\begin{figure}[tbh!]
    \centering
    {\begin{tikzpicture}
    \pgfplotstableread[col sep=comma]{raw_data/rbmqv1/gazebo/outputs_2000_50_run1.csv}\datatable
    \pgfplotstableread[col sep=comma]{raw_data/rbmqv1/gazebo/outputs_2000_50_run2.csv}\datatabletwo
    \pgfplotstableread[col sep=comma]{raw_data/rbmqv1/gazebo/outputs_2000_50_run3.csv}\datatablethree
    \pgfplotstableread[col sep=comma]{raw_data/rbmqv1/gazebo/outputs_2000_50_run4.csv}\datatablefour
    \pgfplotstableread[col sep=comma]{raw_data/rbmqv1/gazebo/outputs_2000_50_run5.csv}\datatablefive
    \pgfplotstablegetrowsof{\datatable}
    \pgfmathsetmacro{\N}{\pgfplotsretval}  
    \begin{groupplot}[group style={
                    group name=myplot,
                    group size= 2 by 3,horizontal sep=.8cm},height=3.5cm,width=4.5cm,
                    minor x tick num = 4,
                    %minor y tick num = 4,
                    %xminorgrids=true,
                    %yminorgrids=true,
                    %xmajorgrids=true,
                    %ymajorgrids=true,
                    legend style={at={(2.7,1)},anchor=north,legend columns=1},
                    ]
        \nextgroupplot[title={Runs $1-5$},ylabel={Sequence numbers}, xlabel={sim-time (s)}]

            %\addplot[no markers, blue] table [x index = {0}, y index = {5},skip coords between index={21}{\N}] \datatable;
            \addplot[no markers, blue, dash dot,thick] table [x index = {0}, y index = {5}] \datatable;
            \addplot[no markers,red,dashed,thick] table [x index = {0}, y index = {5}] \datatabletwo;
            \addplot[no markers, green,thick] table [x index = {0}, y index = {5}] \datatablethree;
            \addplot[no markers,cyan,dashed,thick] table [x index = {0}, y index = {5}] \datatablefour;
            \addplot[no markers, magenta,thick] table [x index = {0}, y index = {5}] \datatablefive;

        \addlegendentry{{run $1$}}
        \addlegendentry{{run $2$}}
        \addlegendentry{{run $3$}}
        \addlegendentry{{run $4$}}
        \addlegendentry{{run $5$}}
        
        \nextgroupplot[title={Runs $1-2$}, xlabel={sim-time (s)}]

        %\addplot[no markers, blue] table [x index = {0}, y index = {5},skip coords between index={21}{\N}] \datatable;
        \addplot[no markers, blue, dash dot,thick] table [x index = {0}, y index = {5}] \datatable;
        \addplot[no markers,red,dashed,thick] table [x index = {0}, y index = {5}] \datatabletwo;

        \nextgroupplot[
            axis y line=none,
            axis x line*=bottom,
            grid=none,
            xlabel=Run 1: real-time (s),
            xtick = {0,0,3.021,8.092},
            xticklabels = {$0$,$0$,$3.021$,$8.092$},
            height=2.3cm,
            enlargelimits=true,
            yshift=10pt
        ]
        \nextgroupplot[
            axis y line=none,
            axis x line*=bottom,
            grid=none,
            xlabel=Run 1: real-time (s),
            xtick = {0,0,3.021,8.092},
            xticklabels = {$0$,$0$,$3.021$,$8.092$},
            height=2.3cm,
            enlargelimits=true,
            yshift=10pt
        ]
        \nextgroupplot[
            axis y line=none,
            axis x line*=bottom,
            grid=none,
            xlabel=Run 2: real-time (s),
            xtick = {0,0,5.433,8.837},
            xticklabels = {$0$,$0$,$5.433$,$8.837$},
            height=2.3cm,
            enlargelimits=true,
            yshift=10pt
        ]
        \nextgroupplot[
            axis y line=none,
            axis x line*=bottom,
            grid=none,
            xlabel=Run 2: real-time (s),
            xtick = {0,0,5.433,8.837},
            xticklabels = {$0$,$0$,$5.433$,$8.837$},
            height=2.3cm,
            enlargelimits=true,
            yshift=10pt
        ]
        
    \end{groupplot}
\end{tikzpicture}}
    
    \caption{Messages outputted by the RMQFMU\textsubscript{1} at every step, over 5 runs, for \textit{maxage} equal to $2000$ms and \textit{lookahead} equal to $50$.}
    \label{fig:fig2}
\end{figure}

%\gita{have some transition sentences that prepare the scene for phase 2, considering that the motivation for this was to remove the maxage effect, and it doesn't come up immediately when we discuss case-study 1 first in phase 2.}
%\henrik{Perhaps, what I wrote as introduction in the following section is actually enough to transition to that part?}
\subsection{Phase 2 Results}
As discussed in the previous results, the RMQFMU\textsubscript{0,1} has a rather conservative interpretation of the \textit{maxage} parameter. 
The FMU will \textit{always} stay at its current output, if the timestamp of the output is   within \textit{maxage} range from the current simulation step timestamp. 
This means that, the timestamp of the output plus the \textit{maxage} is greater than or equal to the current simulation timestamp. 
However, this interpretation is rather strict, in the sense that it doesn't take into account if newer data is present in the incoming FMU queue. 
Thus, we want to relax this condition such that the FMU will only stay at its current output, if the output is within the        \textit{maxage} and in addition there is no newer data available. 
In other words, the FMU should always move ahead to the latest input data valid within the current simulation step, and the \textit{maxage} should only be considered when there is no new input.

We have implemented this change of \textit{maxage} semantics in the RMQFMU\textsubscript{2} version of the FMU. 
The change itself had some side-effects impacting the performance of the FMU step function. 
Without going into details, the change means that the FMU now needs to consider if data is present before considering the maxage. 
However, situations may occur where data is present in the incoming queue, but it may have time-stamps in the future of the current simulation time. This can occur e.g.\ when replaying with a fixed interval historical data including gaps. 
If this situation occurs, the RMQFMU\textsubscript{0,1} versions of the FMU would move the future data from the incoming queue   into an internal processing queue, that would gradually keep increasing in size and thereby impacting the step performance. 
To remedy this issue, RMQFMU\textsubscript{2} has an additional change to the internal queue structure.
In the following paragraphs we present results of experiments with our two case-studies using RMQFMU\textsubscript{2}.

\subsubsection{Case-study 1}
The effect of the RMQFMU\textsubscript{2} changes to the case-study 1 can be observed in Figure~\ref{fig:fig5}. The test configuration for this test is identical to Case 4 of Table~\ref{tab:exp1}. The results of this case with RMQFMU\textsubscript{1} are present in Figure~\ref{fig:log_f2_s100_z100_orig}.
%\henrik{We should keep only f2-s100-z100 results in the Figure~\ref{fig:fig5}. These are sufficient to demonstrate the effects.}
To demonstrate the effect of moving to the latest input data available, we have specified a lookahead size of $100$ in this test. 
This covers more messages than being produced on average within a simulation time step of $100$ms and a   data frequency of $2$ms, which is approximately $100 / 2 = 50$ messages. 
So it basically means consuming as many messages as possible within a time step.

% It can be observed from Figure~\ref{fig:fig5} middle graph, that now within the initial \textit{maxage} period of $300$ms, the seqno output actually increases.
\ifdefined\TECHREP

The middle graph in Figure~\ref{fig:fig5} illustrates,
that the seqno output increases within the initial \textit{maxage} period of $300$ms. This, in contrast to the initial $300$ms period in the middle graph of Figure~\ref{fig:log_f2_s100_z100_orig} where the output stays unchanged. 
\fi
It is possible to observe from Figure~\ref{fig:fig5}, 
\ifdefined\PAPER
graph on the left, 
\fi
\ifdefined\TECHREP
top graph, 
\fi
that in contrast to Figure~\ref{fig:log_f2_s100_z100_orig}, the step duration now has decreased from around $6000$us to around $1000$us. 
This is due to the fact that we now use the larger (100 = move as close to step time as possible) \textit{lookahead} size as opposed to the original size 1, additionally to the improvement in the internal queue implementation. Also, the larger \textit{lookahead} size has the effect, that the queue size (right-hand side) stays low for a longer simulation time compared to earlier test results.
But still the queue size starts increasing due to both data being produced more frequently than consumed and to gaps in the input data. Thus, as mentioned before, the FMU needs to guard any internal queue size limitations. 

\ifdefined\TECHREP
\begin{figure}[tbh!]
    \centering
    {\begin{tikzpicture}
    \pgfplotstableread[col sep=comma]{raw_data/rbmqv2/urdata/cton_f2_s100_z100.csv}\datatable
    \pgfplotstableread[col sep=comma]{raw_data/rbmqv2/urdata/cton_f2_s100_z120.csv}\datatabletwo
    \pgfplotstableread[col sep=comma]{raw_data/rbmqv2/urdata/cton_f2_s2_z2_cut.csv}\datatablethree
    %\pgfplotstablegetrowsof{\datatable}
    %\pgfmathsetmacro{\N}{\pgfplotsretval} 
    \begin{groupplot}[group style={
                    group name=myplot,
                    group size= 1 by 3,vertical sep=1cm},height=4cm,width=8cm,
                    legend style={at={(0.5,-3.3)},anchor=north,legend columns=-1},
                    minor x tick num = 4,
                    %minor y tick num = 4,
                    %xminorgrids=true,
                    %yminorgrids=true,
                    %xmajorgrids=true,
                    %ymajorgrids=true,
                    scaled y ticks=base 10:-3,
                    ]
        \nextgroupplot[ylabel={Step duration $\mu s$}]
            \addplot[mark=none, blue,thick] table[x index = {0},y index = {1}] \datatable;
            %\addplot[mark=none, red, dashed,thick] table[x index = {0},y index = {1}] \datatabletwo;
            %\addplot[mark=none, cyan, dash dot,thick] table[x index = {0},y index = {1}] \datatablethree;

            %\addlegendentry{{$s100\_z100$}}
            %\addlegendentry{{$s100\_z120$}}
            %\addlegendentry{{$\_s2\_z2$}}
            
            \addplot [name path=upper,draw=none] table[x index = {0},y expr=\thisrowno{1}+\thisrowno{2}] \datatable;
            \addplot [name path=lower,draw=none] table[x index = {0},y expr=\thisrowno{1}-\thisrowno{2}] \datatable;
            \addplot [fill=blue!20] fill between[of=upper and lower];
            
            %\addplot [name path=upper,draw=none] table[x index = {0},y expr=\thisrowno{1}+\thisrowno{2}] \datatabletwo;
            %\addplot [name path=lower,draw=none] table[x index = {0},y expr=\thisrowno{1}-\thisrowno{2}] \datatabletwo;
            %\addplot [fill=red!20] fill between[of=upper and lower];
            
            %\addplot [name path=upper,draw=none] table[x index = {0},y expr=\thisrowno{1}+\thisrowno{2}] \datatablethree;
            %\addplot [name path=lower,draw=none] table[x index = {0},y expr=\thisrowno{1}-\thisrowno{2}] \datatablethree;
            %\addplot [fill=cyan!20] fill between[of=upper and lower];

        \nextgroupplot[ylabel={Sequence no}]
            \addplot[mark=none, blue, thick] table[x index = {0},y index = {3}] \datatable;
            %\addplot[mark=none, red, dashed,thick] table[x index = {0},y index = {3}] \datatabletwo;
            %\addplot[mark=none, cyan,dash dot,thick] table[x index = {0},y index = {3}] \datatablethree;

            \addplot [name path=upper,draw=none] table[x index = {0},y expr=\thisrowno{3}+\thisrowno{4}] \datatable;
            \addplot [name path=lower,draw=none] table[x index = {0},y expr=\thisrowno{3}-\thisrowno{4}] \datatable;
            \addplot [fill=blue!20] fill between[of=upper and lower];
            
            %\addplot [name path=upper,draw=none] table[x index = {0},y expr=\thisrowno{3}+\thisrowno{4}] \datatabletwo;
            %\addplot [name path=lower,draw=none] table[x index = {0},y expr=\thisrowno{3}-\thisrowno{4}] \datatabletwo;
            %\addplot [fill=red!20] fill between[of=upper and lower];
            
            %\addplot [name path=upper,draw=none] table[x index = {0},y expr=\thisrowno{3}+\thisrowno{4}] \datatablethree;
            %\addplot [name path=lower,draw=none] table[x index = {0},y expr=\thisrowno{3}-\thisrowno{4}] \datatablethree;
            %\addplot [fill=cyan!20] fill between[of=upper and lower];
            
        \nextgroupplot[ylabel={Queue size}, xlabel={sim-time}]
            \addplot[mark=none, blue,thick] table[x index = {0},y index = {5}] \datatable;
            %\addplot[mark=none, red, dashed,thick] table[x index = {0},y index = {5}] \datatabletwo;
            %\addplot[mark=none, cyan, dash dot,thick] table[x index = {0},y index = {5}] \datatablethree;

            \addplot [name path=upper,draw=none] table[x index = {0},y expr=\thisrowno{5}+\thisrowno{6}] \datatable;
            \addplot [name path=lower,draw=none] table[x index = {0},y expr=\thisrowno{5}-\thisrowno{6}] \datatable;
            \addplot [fill=blue!20] fill between[of=upper and lower];
            
            %\addplot [name path=upper,draw=none] table[x index = {0},y expr=\thisrowno{5}+\thisrowno{6}] \datatabletwo;
            %\addplot [name path=lower,draw=none] table[x index = {0},y expr=\thisrowno{5}-\thisrowno{6}] \datatabletwo;
            %\addplot [fill=red!20] fill between[of=upper and lower];
            
            %\addplot [name path=upper,draw=none] table[x index = {0},y expr=\thisrowno{5}+\thisrowno{6}] \datatablethree;
            %\addplot [name path=lower,draw=none] table[x index = {0},y expr=\thisrowno{5}-\thisrowno{6}] \datatablethree;
            %\addplot [fill=cyan!20] fill between[of=upper and lower];

    \end{groupplot}

\end{tikzpicture}}
    
    \caption{Step duration and sequence numbers outputted by RMQFMU\textsubscript{2} at every step, for $t_{\mathit{on}}$, and $f_{data}=2$ms.}
    \label{fig:fig5}
\end{figure}
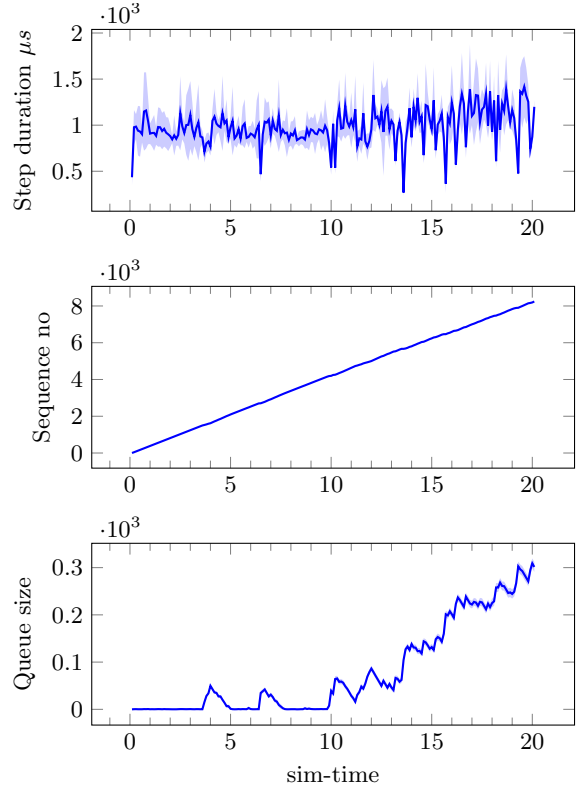
\fi
\ifdefined\PAPER
\begin{figure}[tbh!]
    \centering
    {\begin{tikzpicture}
    \pgfplotstableread[col sep=comma]{raw_data/rbmqv2/urdata/cton_f2_s100_z100.csv}\datatable
    \pgfplotstableread[col sep=comma]{raw_data/rbmqv2/urdata/cton_f2_s100_z120.csv}\datatabletwo
    \pgfplotstableread[col sep=comma]{raw_data/rbmqv2/urdata/cton_f2_s2_z2_cut.csv}\datatablethree
    \pgfplotstablegetrowsof{\datatable}
    \pgfmathsetmacro{\N}{\pgfplotsretval} 
    \begin{groupplot}[group style={
                    group name=myplot,
                    group size= 2 by 1,vertical sep=1.cm},height=3.5cm,width=6cm,
                    legend style={at={(0.5,-.2)},anchor=north,legend columns=-1},
                    minor x tick num = 4,
                    %minor y tick num = 1,
                    %xminorgrids=true,
                    %yminorgrids=true,
                    %xmajorgrids=true,
                    %ymajorgrids=true,
                    ymin=0,
                    xmax=21,
                    scaled y ticks=base 10:-3,
                    ]
        \nextgroupplot[ylabel={Step duration ($\mu s$}), xlabel={sim-time (s)}]
            \addplot[mark=none, blue] table[x index = {0},y index = {1}] \datatable;
            %\addplot[mark=none, red, dashed,thick] table[x index = {0},y index = {1}] \datatabletwo;
            %\addplot[mark=none, cyan, dash dot,thick] table[x index = {0},y index = {1}] \datatablethree;

            %\addlegendentry{{$t_{\mathit{on}}\_f2\_s100\_z100$}}
            %\addlegendentry{{$t_{on}\_f2\_s100\_z120$}}
            %\addlegendentry{{$t_{on}\_f2\_s2\_z2$}}
            
            \addplot [name path=upper,draw=none] table[x index = {0},y expr=\thisrowno{1}+\thisrowno{2}] \datatable;
            \addplot [name path=lower,draw=none] table[x index = {0},y expr=\thisrowno{1}-\thisrowno{2}] \datatable;
            \addplot [fill=blue!20] fill between[of=upper and lower];

        \nextgroupplot[ylabel={Queue size},xlabel={sim-time (s)},scaled y ticks=base 10:-2]
            \addplot[mark=none, blue] table[x index = {0},y index = {5}] \datatable;
            %\addplot[mark=none, red, dashed,thick] table[x index = {0},y index = {5}] \datatabletwo;
            %\addplot[mark=none, cyan, dash dot,thick] table[x index = {0},y index = {5}] \datatablethree;

            \addplot [name path=upper,draw=none] table[x index = {0},y expr=\thisrowno{5}+\thisrowno{6}] \datatable;
            \addplot [name path=lower,draw=none] table[x index = {0},y expr=\thisrowno{5}-\thisrowno{6}] \datatable;
            \addplot [fill=blue!20] fill between[of=upper and lower];
            
            %\addplot [name path=upper,draw=none] table[x index = {0},y expr=\thisrowno{5}+\thisrowno{6}] \datatabletwo;
            %\addplot [name path=lower,draw=none] table[x index = {0},y expr=\thisrowno{5}-\thisrowno{6}] \datatabletwo;
            %\addplot [fill=red!20] fill between[of=upper and lower];
            
            %\addplot [name path=upper,draw=none] table[x index = {0},y expr=\thisrowno{5}+\thisrowno{6}] \datatablethree;
            %\addplot [name path=lower,draw=none] table[x index = {0},y expr=\thisrowno{5}-\thisrowno{6}] \datatablethree;
            %\addplot [fill=cyan!20] fill between[of=upper and lower];
    \end{groupplot}

    % this is the inset plot ...
    %\begin{axis}[
    %	tiny,
        % ... which should be plotted at the stored coordinate ...
    %    at={(pt)},
        % ... with this `anchor'
    %    anchor=north east,
        % use this predefined style (it is predefined by PGFPlots itself)
        %
        % now state the options which should be used for the inset plot
    %    width=2.5cm,
    %    height=2.5cm,
    %    xtick distance=.2,
    %    xmin=0,xmax=.4,
        %xmajorgrids=true,
        %ymajorgrids=true,
        % use this key to fill the background of the axis only
    %    axis background/.style={
    %        fill=white,
    %    },
        % name this axis so it can later be used to fill the "background" of the
        % whole plot including the labels
    %    name=insetAxis,
    %]
%        \addplot [mark=x] table [x=size, y=cluster] {sim-0.3-0.6.csv};
        % again the dummy plot
        
    %    \addplot[mark=none, blue] table[x index = {0},y index = {3},skip coords between index={5}{\N}] \datatable;

    %\end{axis}
%
\end{tikzpicture}}
    
    \caption{Step duration and sequence numbers outputted by RMQFMU\textsubscript{2} at every step, for step-size of $100$ms, and delay of $100$ms.}
    \label{fig:fig5}
\end{figure}
\fi

%In the following section we elaborate further on the inter-dependency between the \textit{maxage} and \textit{looakahead size} for the RMQFMU\textsubscript{2} in the context of case-study 2.

\subsubsection{Case-study 2}
The new version of the RMQFMU\textsubscript{2} removes the delay in the outputted sequence numbers noticeable for large values of the \textit{maxage}, e.g. $2000$ms (Figure~\ref{fig:fig6}, for Cases $1-4$), as compared to RMQFMU\textsubscript{1} (Figure~\ref{fig:fig4}). 
The utility of the \textit{maxage} parameter becomes evident, once gaps are introduced in the data, in these experiments equal to $500$ms, Figure~\ref{fig:fig6b}, left graph, and $1000$ms, right graph.
It can be observed how when there is no input, the sequence value that is outputted is the last value that is valid from a time perspective. 
Naturally, for larger gaps, the sequence numbers outputted are more delayed, e.g. for $t=5$s, the RMQFMU\textsubscript{2} with $500$ms gaps is at number $10$, whereas for $1000$ms is at number $5$.
\ifdefined\TECHREP
Note, for the bottom graphs the \textit{lookahead} is set to $10$ instead of $50$, however, that will not have an impact in the results, as with the given gaps, there cannot be more than $2$ and $1$ messages respectively (for $500$ms and $1000$ms delays) at a time in the queue.
\fi
\ifdefined\PAPER
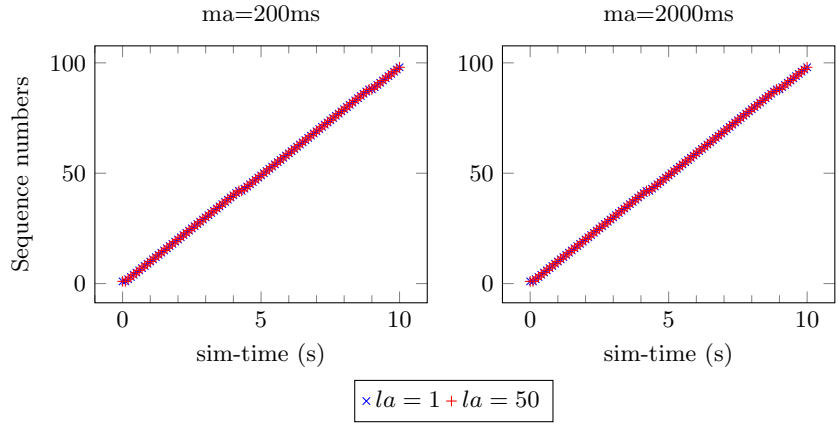
\begin{figure}[bth!]
    \centering
    {\begin{tikzpicture}
    \pgfplotstableread[col sep=comma]{raw_data/rbmqv2/gazebo/outputs_ma200_la1.csv}\datatable
    \pgfplotstableread[col sep=comma]{raw_data/rbmqv2/gazebo/outputs_ma200_la50.csv}\datatabletwo
    \pgfplotstableread[col sep=comma]{raw_data/rbmqv2/gazebo/outputs_ma2000_la1.csv}\datatablethree
    \pgfplotstableread[col sep=comma]{raw_data/rbmqv2/gazebo/outputs_ma2000_la50.csv}\datatablefour
    \begin{groupplot}[group style={
                    group name=myplot,
                    group size= 2 by 1,horizontal sep=.8cm},height=3.5cm,width=4.5cm,
                    legend style={nodes={scale=0.7, transform shape},at={(2.7,1.0)},anchor=north,legend columns=1},
                    minor x tick num = 4,
                    minor y tick num = 4,
                    %xminorgrids=true,
                    %yminorgrids=true,
                    %xmajorgrids=true,
                    %ymajorgrids=true,
                    ]
        \nextgroupplot[title={ma=$200$ms},ylabel={Sequence numbers}, xlabel={sim-time (s)}]
            \addplot[no marks, blue, very thick, dash dot] table [x index = {0}, y index = {5}] \datatable;
            \addplot[no marks, red, dashed, very thick] table [x index = {0}, y index = {5}] \datatabletwo;
        \addlegendentry{{$la=1$}}
        \addlegendentry{{$la=50$}}
        \nextgroupplot[title={ma=$2000$ms}, xlabel={sim-time (s)}]
            \addplot[no marks, blue, very thick, dash dot] table [x index = {0}, y index = {5}] \datatablethree;
            \addplot[no marks, red, dashed, very thick] table [x index = {0}, y index = {5}] \datatablefour;

    \end{groupplot}
    %\node (title) at ($(myplot c1r1.center)!0.5!(myplot c2r1.center)+(0,4cm)$) {Sequence numbers over time for $f_{data}=100$ms};
\end{tikzpicture}}
    
    \caption{Messages outputted by the RMQFMU\textsubscript{2} at every step, for $f_{data}=100$ms.}
    \label{fig:fig6}
\end{figure}
\begin{figure}[tbh!]
    \centering
    {\begin{tikzpicture}
    \pgfplotstableread[col sep=comma]{raw_data/rbmqv2/gazebo/outputs_ma2000_la1_500msgap.csv}\datatablefive
    \pgfplotstableread[col sep=comma]{raw_data/rbmqv2/gazebo/outputs_ma2000_la10_500msgap.csv}\datatablesix
    \pgfplotstableread[col sep=comma]{raw_data/rbmqv2/gazebo/outputs_ma2000_la1_1000msgap.csv}\datatableseven
    \pgfplotstableread[col sep=comma]{raw_data/rbmqv2/gazebo/outputs_ma2000_la10_1000msgap.csv}\datatableeight
    \begin{groupplot}[group style={
                    group name=myplot,
                    group size= 2 by 1,horizontal sep=.8cm},height=3.5cm,width=4.5cm,
                    legend style={nodes={scale=0.7, transform shape},at={(2.7,1.0)},anchor=north,legend columns=1},
                    minor x tick num = 4,
                    %minor y tick num = 4,
                    xmin=-0.2,
                    ymin=0,
                    ytick distance = 5,
                    %xminorgrids=true,
                    %yminorgrids=true,
                    %xmajorgrids=true,
                    %ymajorgrids=true,
                    ]

        \nextgroupplot[title={ma=$2000$ms},ylabel={Sequence numbers},xlabel={sim-time (s)}]
            \addplot[no marks, blue] table [x index = {0}, y index = {5}] \datatablefive;
            %\addplot[no marks, very thick, dashed,red] table [x index = {0}, y index = {5}] \datatablesix;

        %\addlegendentry{{$la=1$}}
        %\addlegendentry{{$la=10$}}
        \nextgroupplot[title={ma=$2000$ms},xlabel={sim-time (s)}]
            \addplot[no marks, blue] table [x index = {0}, y index = {5}] \datatableseven;
            %\addplot[no marks, very thick, dashed,red] table [x index = {0}, y index = {5}] \datatableeight;

    \end{groupplot}
    %\node (title) at ($(myplot c1r1.center)!0.5!(myplot c2r1.center)+(0,4cm)$) {Sequence numbers over time for $f_{data}=100$ms};
\end{tikzpicture}}
    
    \caption{Messages outputted by the RMQFMU\textsubscript{2} at every step, for $f_{data}=100$ms and \textit{lookahead}$=1$, with gaps of $500$ms (left), and $1000$ms (right).}
    \label{fig:fig6b}
\end{figure}
\fi
\ifdefined\TECHREP
\begin{figure}[tbh!]
    \centering
    {\begin{tikzpicture}
    \pgfplotstableread[col sep=comma]{raw_data/rbmqv2/gazebo/outputs_ma200_la1.csv}\datatable
    \pgfplotstableread[col sep=comma]{raw_data/rbmqv2/gazebo/outputs_ma200_la50.csv}\datatabletwo
    \pgfplotstableread[col sep=comma]{raw_data/rbmqv2/gazebo/outputs_ma2000_la1.csv}\datatablethree
    \pgfplotstableread[col sep=comma]{raw_data/rbmqv2/gazebo/outputs_ma2000_la50.csv}\datatablefour
    \begin{groupplot}[group style={
                    group name=myplot,
                    group size= 2 by 1,vertical sep=1.cm},
                    height=5cm, width=6cm,
                    legend style={at={(1.08,-0.3)},anchor=north,legend columns=-1},
                    minor x tick num = 4,
                    %minor y tick num = 4,
                    %xminorgrids=true,
                    %yminorgrids=true,
                    %xmajorgrids=true,
                    %ymajorgrids=true,
                    ]
        \nextgroupplot[title={ma=$200$ms},ylabel={Sequence numbers}, xlabel={sim-time (s)}]
            \addplot[only marks, mark=x, blue] table [x index = {0}, y index = {5}] \datatable;
            \addplot[only marks, mark=+,red] table [x index = {0}, y index = {5}] \datatabletwo;
        \addlegendentry{{$la=1$}}
        \addlegendentry{{$la=50$}}
        \nextgroupplot[title={ma=$2000$ms}, xlabel={sim-time (s)}]
            \addplot[only marks, mark=x, blue] table [x index = {0}, y index = {5}] \datatablethree;
            \addplot[only marks, mark=+,red] table [x index = {0}, y index = {5}] \datatablefour;

    \end{groupplot}
    %\node (title) at ($(myplot c1r1.center)!0.5!(myplot c2r1.center)+(0,4cm)$) {Sequence numbers over time for $f_{data}=100$ms};
\end{tikzpicture}}
    
    \caption{Messages outputted by the RMQFMU\textsubscript{2} at every step, for $f_{data}=100$ms.}
    \label{fig:fig6}
\end{figure}
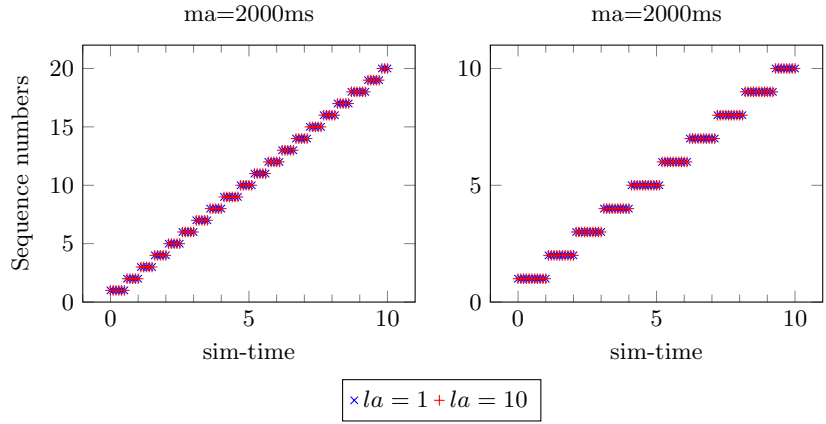
\begin{figure}[tbh!]
    \centering
    {\begin{tikzpicture}
    \pgfplotstableread[col sep=comma]{raw_data/rbmqv2/gazebo/outputs_ma2000_la1_500msgap.csv}\datatablefive
    \pgfplotstableread[col sep=comma]{raw_data/rbmqv2/gazebo/outputs_ma2000_la10_500msgap.csv}\datatablesix
    \pgfplotstableread[col sep=comma]{raw_data/rbmqv2/gazebo/outputs_ma2000_la1_1000msgap.csv}\datatableseven
    \pgfplotstableread[col sep=comma]{raw_data/rbmqv2/gazebo/outputs_ma2000_la10_1000msgap.csv}\datatableeight
    \begin{groupplot}[group style={
                    group name=myplot,
                    group size= 2 by 1,vertical sep=1.cm},
                    height=5cm, width=6cm,
                    legend style={at={(1.08,-0.3)},anchor=north,legend columns=-1},
                    minor x tick num = 4,
                    %minor y tick num = 4,
                    ymin=0,
                    ytick distance = 5,
                    %xminorgrids=true,
                    %yminorgrids=true,
                    %xmajorgrids=true,
                    %ymajorgrids=true,
                    ]

        \nextgroupplot[title={ma=$2000$ms},ylabel={Sequence numbers},xlabel={sim-time}]
            \addplot[only marks, mark=x, blue] table [x index = {0}, y index = {5}] \datatablefive;
            \addplot[only marks, mark=+,red] table [x index = {0}, y index = {5}] \datatablesix;

        \addlegendentry{{$la=1$}}
        \addlegendentry{{$la=10$}}
        \nextgroupplot[title={ma=$2000$ms},xlabel={sim-time}]
            \addplot[only marks, mark=x, blue] table [x index = {0}, y index = {5}] \datatableseven;
            \addplot[only marks, mark=+,red] table [x index = {0}, y index = {5}] \datatableeight;

    \end{groupplot}
    %\node (title) at ($(myplot c1r1.center)!0.5!(myplot c2r1.center)+(0,4cm)$) {Sequence numbers over time for $f_{data}=100$ms};
\end{tikzpicture}}
    
    \caption{Messages outputted by the RMQFMU\textsubscript{2} at every step, for $f_{data}=100$ms with gaps of $500$ms (left), and $1000$ms (right).}
    \label{fig:fig6b}
\end{figure}
\fi

In order to show the utility of the \textit{lookahead}, we ran the experiments with simulation step-size equal to $100$ms, for values of \textit{maxage} in \{200ms, 2000ms\}, with frequency of sending data equal to $500$Hz (Figure~\ref{fig:fig7}). 
For smaller values of the \textit{maxage}, the delay of the RMQFMU\textsubscript{2} for \textit{lookahead} equal to $1$, while consistently present, is rather low.
However, for larger values of the \textit{maxage}, the delay for low \textit{lookahead} values is non-negligible.
It is possible to observe that for the duration of the \textit{maxage} the sequence numbers gradually increase by $1$, as expected for \textit{lookahead} equal to $1$.
However, after the initial \textit{maxage} duration has passed, there is a jump in the sequence numbers, that continue to follow the results for \textit{lookahead} equal to $50$, albeit with a more or less constant delay until the end.
A similar effect was observed in Figure~\ref{fig:fig4}, with difference that in the current version, the RMQFMU\textsubscript{2} will check if there is a newer message which it can output.
Such effect is due to the fact that after $2$s, older in-between values become invalid, and thus are not outputted, with RMQFMU\textsubscript{2} going back to the queue to pick up more recent messages. 

\ifdefined\TECHREP
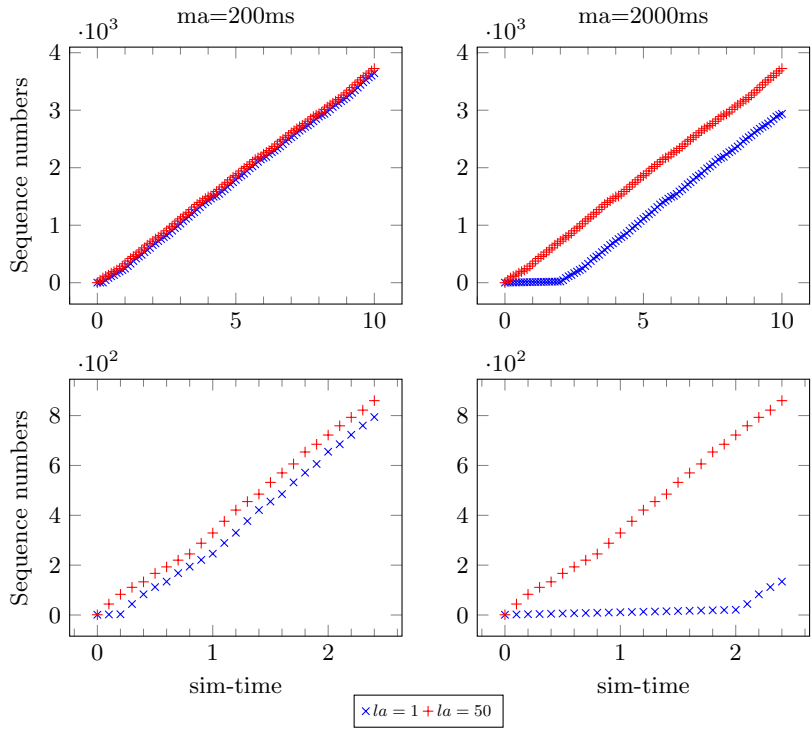
\begin{figure}[tbh!]
    \centering
    {\begin{tikzpicture}
    \pgfplotstableread[col sep=comma]{raw_data/rbmqv2/gazebo/outputs_ma200_la1_f2ms.csv}\datatable
    \pgfplotstableread[col sep=comma]{raw_data/rbmqv2/gazebo/outputs_ma200_la50_f2ms.csv}\datatabletwo
    \pgfplotstableread[col sep=comma]{raw_data/rbmqv2/gazebo/outputs_ma2000_la1_f2ms.csv}\datatablethree
    \pgfplotstableread[col sep=comma]{raw_data/rbmqv2/gazebo/outputs_ma2000_la50_f2ms.csv}\datatablefour
    \pgfplotstablegetrowsof{\datatable}
    \pgfmathsetmacro{\N}{\pgfplotsretval} 
    \begin{groupplot}[group style={
                    group name=myplot,
                    group size= 2 by 2,vertical sep=1cm},height=5cm,width=6cm,
                    legend style={nodes={scale=0.7, transform shape}, at={(1.08,-1.52)},anchor=north,legend columns=-1},
                    minor x tick num = 4,
                    %minor y tick num = 4,
                    %xminorgrids=true,
                    %yminorgrids=true,
                    %xmajorgrids=true,
                    %ymajorgrids=true,
                    scaled y ticks=base 10:-3,
                    ]
        \nextgroupplot[title={ma=$200$ms},ylabel={Sequence numbers}]
            \addplot[only marks, mark=x, blue] table [x index = {0}, y index = {5}] \datatable;
            \addplot[only marks, mark=+,red] table [x index = {0}, y index = {5}] \datatabletwo;
        \addlegendentry{{$la=1$}}
        \addlegendentry{{$la=50$}}
        \nextgroupplot[title={ma=$2000$ms}]
            \addplot[only marks, mark=x, blue] table [x index = {0}, y index = {5}] \datatablethree;
            \addplot[only marks, mark=+,red] table [x index = {0}, y index = {5}] \datatablefour;
            
        \nextgroupplot[
                    scaled y ticks=base 10:-2,ylabel={Sequence numbers},xlabel=sim-time]
            \addplot[only marks, mark=x, blue] table [x index = {0}, y index = {5},skip coords between index={25}{\N}] \datatable;
            \addplot[only marks, mark=+,red] table [x index = {0}, y index = {5},skip coords between index={25}{\N}] \datatabletwo;

        \nextgroupplot[
                    scaled y ticks=base 10:-2,xlabel=sim-time]
            \addplot[only marks, mark=x, blue] table [x index = {0}, y index = {5},skip coords between index={25}{\N}] \datatablethree;
            \addplot[only marks, mark=+,red] table [x index = {0}, y index = {5},skip coords between index={25}{\N}] \datatablefour;
    \end{groupplot}
    %\node (title) at ($(myplot c1r1.center)!0.5!(myplot c2r1.center)+(0,25cm)$) {Sequence numbers over time for $f_{data}=2$ms};
\end{tikzpicture}}
    
    \caption{Messages outputted by the RMQFMU\textsubscript{2} at every step, for $f_{data}=2$ms, over $10$s (top graphs), zoomed in from $0-2$s (bottom graphs).}
    \label{fig:fig7}
\end{figure}
\fi

\ifdefined\PAPER
\begin{figure}[tbh!]
    \centering
    \resizebox{\textwidth}{!}{\pgfdeclarelayer{background}
\pgfdeclarelayer{foreground}
\pgfsetlayers{background,main,foreground}
\begin{tikzpicture}
    \pgfplotstableread[col sep=comma]{raw_data/rbmqv2/gazebo/outputs_ma200_la1_f2ms.csv}\datatable
    \pgfplotstableread[col sep=comma]{raw_data/rbmqv2/gazebo/outputs_ma200_la50_f2ms.csv}\datatabletwo
    \pgfplotstableread[col sep=comma]{raw_data/rbmqv2/gazebo/outputs_ma2000_la1_f2ms.csv}\datatablethree
    \pgfplotstableread[col sep=comma]{raw_data/rbmqv2/gazebo/outputs_ma2000_la50_f2ms.csv}\datatablefour
    \pgfplotstablegetrowsof{\datatable}
    \pgfmathsetmacro{\N}{\pgfplotsretval} 
    \begin{pgfonlayer}{background}
    \begin{groupplot}[group style={
                    group name=myplot,
                    group size= 2 by 1,horizontal sep=.8cm},height=4cm,width=5cm,
                    legend style={nodes={scale=0.7, transform shape}, at={(2.6,1.0)},anchor=north,legend columns=1},
                    minor x tick num = 4,
                    %minor y tick num = 4,
                    %xminorgrids=true,
                    %yminorgrids=true,
                    %xmajorgrids=true,
                    %ymajorgrids=true,
                    scaled y ticks=base 10:-3,
                    xmax=11
                    ]
            
        \nextgroupplot[ylabel={Sequence numbers},xlabel={sim-time (s)}]
            \addplot[no marks, blue, very thick, dash dot] table [x index = {0}, y index = {5}] \datatable;
            \addplot[no marks, very thick, dotted,red] table [x index = {0}, y index = {5}] \datatabletwo;
\coordinate (pt0) at (axis cs:11.7,2200);
\draw[->, >=stealth', black!70, dashed] (axis cs:1,70) -- (axis cs:7.4,800);
        \addlegendentry{{$la=1$}}
        \addlegendentry{{$la=50$}}
        \nextgroupplot[xlabel={sim-time (s)}]
            %\addplot[only marks, mark=x, blue] table [x index = {0}, y index = {5}] \datatablethree;
            %\addplot[only marks, mark=+,red] table [x index = {0}, y index = {5}] \datatablefour;
            \addplot[no marks, blue, very thick, dash dot] table [x index = {0}, y index = {5}] \datatablethree;
            \addplot[no marks, very thick, dashed,red] table [x index = {0}, y index = {5}] \datatablefour;

    \coordinate (pt) at (axis cs:12,2100);
    
    \draw[->, >=stealth', black!70, dashed] (axis cs:3,50) -- (axis cs:7.8,800);
    \end{groupplot}

    \end{pgfonlayer}
    
    \begin{pgfonlayer}{foreground}
    % this is the inset plot ...
    \begin{axis}[
    	tiny,
        % ... which should be plotted at the stored coordinate ...
        at={(pt0)},
        % ... with this `anchor'
        anchor=north east,
        % use this predefined style (it is predefined by PGFPlots itself)
        %
        % now state the options which should be used for the inset plot
        width=2.5cm,
        height=2.5cm,
        xmin=0,xmax=0.2,
        xtick distance=0.1,
        %xmajorgrids=true,
        %ymajorgrids=true,
        % use this key to fill the background of the axis only
        axis background/.style={
            fill=white,
        },
        % name this axis so it can later be used to fill the "background" of the
        % whole plot including the labels
        name=insetAxis,
    ]
%        \addplot [mark=x] table [x=size, y=cluster] {sim-0.3-0.6.csv};
        % again the dummy plot
        \addplot[no marks, blue, very thick, dash dot] table [x index = {0}, y index = {5},skip coords between index={25}{\N}] \datatable;
    \end{axis}
    \end{pgfonlayer}

     \begin{pgfonlayer}{foreground}
    % this is the inset plot ...
    \begin{axis}[
    	tiny,
        % ... which should be plotted at the stored coordinate ...
        at={(pt)},
        % ... with this `anchor'
        anchor=north east,
        % use this predefined style (it is predefined by PGFPlots itself)
        %
        % now state the options which should be used for the inset plot
        width=2.5cm,
        height=2.5cm,
        xmin=0,xmax=2,
        xtick distance=1,
        %xmajorgrids=true,
        %ymajorgrids=true,
        % use this key to fill the background of the axis only
        axis background/.style={
            fill=white,
        },
        % name this axis so it can later be used to fill the "background" of the
        % whole plot including the labels
        name=insetAxis,
    ]
%        \addplot [mark=x] table [x=size, y=cluster] {sim-0.3-0.6.csv};
        % again the dummy plot
        \addplot[no marks, blue, very thick, dash dot] table [x index = {0}, y index = {5},skip coords between index={25}{\N}] \datatablethree;
    \end{axis}
    \end{pgfonlayer}
    
\end{tikzpicture}}
    
    \caption{RMQFMU\textsubscript{2} output at every step, for $f_{data}=2$ms, over $2$s.}
    \label{fig:fig7}
\end{figure}
\fi

%\section{Related Work}
%Check for literature on:
%\begin{itemize}
%    \item we need to look for some other tool that is similar to what the rabbitmq fmu does, otherwise we don't need this section. \gita{agreed.}
%\end{itemize}

\section{RMQFMU\textsubscript{2} Configuration Guidelines}\label{sec:guidelines}
%\begin{itemize}
%    \item Guidelines on how to tune the lookahead and maxage parameters
%    \item Guidelines on when to use the threaded solution
%    \item Other interesting points to discuss?
%    \item relation between lookahead, maxage, step size, frequency of data and RMQFMU outputs (i.e. context of receiving every value)
%\end{itemize}

The parameters of the RMQFMU\textsubscript{2} should be configured based on the use-case in order to result in desired behaviour. 
If possible, the simulation step-size should match the frequency of sending data $f_{data}$.
However, depending on the time granularity needed by different FMUs, achieving such alignment might not be possible. 
The maxage value has to be big enough to make up for any time gap between data, but also low enough to ensure consistency in domain of use. 
E.g. for some applications, it might the case that data should not be older than $200$ms to ensure correct operation.
The lookahead can be set to counter-balance the effect of the maxage, as it allows to jump to the latest data that is available in the queue (within the limits of the lookahead).
Note however, larger values of the lookahead will result in bigger time jumps between the messages outputted.
If more intermittent values are required, then a lower value should be adopted. 
The frequency of sending data influences the meaningful range of values for the lookahead, e.g. sending data every $2$ms, will result in $500$ messages per second.
For a simulation step-size of $100$ms, a reasonable value of the lookahead would be circa $50$.
Finally, the speed of the simulation also affects what is present in the queue at any time, as a result impacting RMQFMU\textsubscript{2} outputs.

%\begin{itemize}

    %\item 
    %\item 
    %Additionally, if it is possible to estimate the the range of potential delays in the transmission of the data, then the lookahead could be set such to cover for that.
%\end{itemize}
%\gita{ADD this: we can have a higher maxage than the freq of sending data, to make up for gaps, or small delays that would get the co-simulation to hang.}
%\gita{ADD this: conclude this subsection with considerations about setting the maxage and lookahead, wrt also frequency of sending data.}

\section{Concluding Remarks}\label{sec:conclusion}
In this paper we have described an extended RMQFMU data-broker that enables coupling an FMI-based co-simulation environment to a non-FMI external component, via the AMQP protocol.
The RMQFMU supports communication in both directions, and can be used in different contexts ranging from replaying historical data into a co-simulation, to enabling the realisation of the DT concept by connecting the DT to its physical counterpart. 
We evaluate this component in terms of its performance, i.e. the real-time duration of a simulation step.
Our results show the benefit of the implementation of a threaded solution, that effectively decouples the \texttt{doStep} logic from the consumption from the rabbitMQ server. 
Moreover, we explore different values of its configurable parameters, such as the \textit{maxage} and \textit{lookahead}, and provide guidelines that can help practitioners in the usage of the RMQFMU. 

There are four main directions for future work. 
First, we are interested in enabling the RMQFMU to take bigger step-sizes if necessary, given that there  is future data to jump to.
In some monitoring applications, where only the latest data-points are relevant, such jump could be of use to mitigation components that deal with out of synchronisation situations, where the DT gets behind in time with respect to the PT.
Second, we plan to formally verify the presented version of the RMQFMU, thus extending previous work that tackled the initial version of the RMQFMU~\cite{thule2020formally}. 
Third, we intend to profile the overhead of RMQFMU as data-broker compared to an FMU that directly queries a database, for example in the context of large amounts of data.
Finally, we will enable the master-algorithm to learn the optimal step-sizes automatically and deal with recurring variation at run-time. This will devoid design-time decisions and adjust step-sizes during run-time as needed.

%\section*{Acknowledgements}
%We are grateful to the Poul Due Jensen Foundation, which has supported the establishment of a new Centre
%for Digital Twin Technology at Aarhus University, which will take forward the principles, tools and
%applications of the engineering of digital twins.

%\input{rbmqFMU_currentDesign}
\typeout{}
\bibliographystyle{unsrt}
\bibliography{rbmq}

\end{document}